\documentclass[journal]{IEEEtran}

\ifCLASSINFOpdf

\usepackage{graphicx}
\graphicspath{ {./images/} }

\else

\fi

\usepackage{xcolor}
\usepackage{amssymb}
\usepackage{cite}
\usepackage{url}
\usepackage{soul}

\begin{document}
\title{A TinyML Platform for On-Device Continual Learning with Quantized Latent Replays}

\author{Leonardo~Ravaglia,
        Manuele~Rusci,
        Davide~Nadalini,
        Alessandro~Capotondi, \\
        Francesco~Conti,~\IEEEmembership{Member,~IEEE},
        Luca~Benini,~\IEEEmembership{Fellow,~IEEE}% <-this % stops a space
\thanks{L. Ravaglia, M. Rusci, D. Nadalini, F. Conti, and L. Benini are with the Department of Electrical, Electronic and Information Engineering (DEI) of the University of Bologna, Viale del Risorgimento 2, 40136 Bologna, Italy (e-mail: \{leonardo.ravaglia2, manuele.rusci, d.nadalini, f.conti, luca.benini\}@unibo.it).}%
\thanks{A. Capotondi is with the Department of Physics, Informatics and Mathematics of the University of Modena and Reggio Emilia, Via Campi 213/A, 41125 Modena, Italy (e-mail: alessandro.capotondi@unibo.it).}%
\thanks{L. Benini is also with the integrated Systems Laboratory (IIS) of ETH Z\"urich, ETZ, Gloriastrasse 35, 8092 Z\"urich, Switzerland (e-mail: lbenini@iis.ee.ethz.ch).}%
\thanks{This   work   was   supported   in   part   by  the ECSEL Horizon 2020 project AI4DI (Artificial intelligence for Digital Industry, g.a. no. 826060); and by EU Horizon 2020 project BonsAPPs (g.a. no. 101015848).
We also acknowledge CINECA for the availability of high-performance computing resources and support awarded under the ISCRA initiative through the NAS4NPC project.
}%
\thanks{Manuscript received May 15, 2021.}}

\maketitle
\begin{abstract}
In the last few years, research and development on Deep Learning models \& techniques for ultra-low-power devices -- in a word, \textit{TinyML} -- has mainly focused on a  \textit{train-then-deploy} assumption, with static models that cannot be adapted to newly collected data without cloud-based data collection and fine-tuning.
Latent Replay-based Continual Learning (CL) techniques\cite{pellegrini2019latent} enable online, serverless adaptation in principle, but so far they have still been too computation- and memory-hungry for ultra-low-power TinyML devices, which are typically based on microcontrollers.
In this work, we introduce a HW/SW platform for end-to-end CL based on a 10-core \textit{FP32}-enabled parallel ultra-low-power (PULP) processor.
We rethink the baseline Latent Replay CL algorithm, leveraging quantization of the frozen stage of the model and Latent Replays (LRs) to reduce their memory cost with minimal impact on accuracy. 
In particular, 8-bit compression of the LR memory proves to be almost lossless (-0.26\% with 3000LR) compared to the full-precision baseline implementation, but requires 4$\times$ less memory, while 7-bit can also be used with an additional minimal accuracy degradation (up to 5\%).
We also introduce optimized primitives for forward and backward propagation on the PULP processor, together with data tiling strategies to fully exploit its memory hierarchy, while maximizing efficiency.
Our results show that by combining these techniques, continual learning can be achieved in practice using less than 64MB of memory -- an amount compatible with embedding in TinyML devices.
On an advanced 22nm prototype of our platform, called \textit{VEGA}, the proposed solution performs on average $65 \times$ faster than a low-power STM32 L4 microcontroller, being $37\times$ more energy efficient -- enough for a lifetime of 535h when learning a new mini-batch of data once every minute.
\end{abstract}


\IEEEpeerreviewmaketitle

\section{Introduction}
The internet-of-Things ecosystem is made possible by miniaturized and smart end-node devices, which can sense the surrounding environment and take decisions based on the information inferred from sensor data. Because of their tiny form factor and the requirement for low cost and battery-operated nature, these smart networked devices are severely constrained in terms of memory capacity and maximum performance and use small Microcontroller Units (MCUs) as their main on-board computing device~\cite{kumar2017resource}. At the same time, there is an ever-growing interest in deploying more accurate and sophisticated data analytics pipelines, such as  Deep Learning (DL) inference models, directly on IoT end-nodes. These competing needs have given rise in the last few years to a specific branch of machine learning (ML) and DL research called \textit{TinyML}~\cite{banbury2020benchmarking} -- focused on shrinking and compressing top-accurate DL models with respect to the target device characteristics.

The primary limitation of the current generation of TinyML hardware and software is that it is mostly focused on \textit{inference}.
The inference task can be strongly optimized by quantizing~\cite{choi2018pact} or pruning~\cite{blalock2020state} the trained model. Many vendors of AI-oriented system-on-chips (SoCs) provide deployment frameworks to automatically translate DL inference graphs into human-readable or machine code~\cite{david2020tensorflow}.
This \textit{train-then-deploy} design process rigidly separates the learning phase from the runtime inference, resulting in a \textit{static} intelligence model design flow, incapable of adapting to phenomena such as \textit{data distribution shift}: a shift in the statistical properties of real incoming data vs the training set that often impacts applications, causing the smart sensors platform to be unreliable when deployed in the field \cite{amodei2016concrete}.

Even if the algorithms themselves are technically capable to learn and adapt to new incoming data, the update process can only be handled from a centralized service, running on the cloud or host servers~\cite{de2021robustifying}.
In this regard, the original training dataset would have to be enriched with the newly collected dataset, and the model would have to be retrained from scratch on the enlarged dataset, adapting to the new data without forgetting the original information~\cite{de2021robustifying}.
Such an adaptive mechanism belongs to the \textit{rehearsal} category and requires the storage of the full training set, often amounting to gigabytes of data.
Additionally, large amounts of data have to be collected in a centralized fashion by network communication, resulting in potential security and privacy concerns, as well as issues of radio power consumption and network reliability in non-urban areas.

{
 
We argue that a robust and privacy-aware solution to these challenges is enabling future smart IoT end-nodes to Lifelong Learning, also known as \textit{Continual Learning}~\cite{song2018situ}(CL): the capability to autonomously adapt to the ever-changing surrounding environment by learning continually (only) from incoming data without forgetting the original knowledge -- a phenomenon known as \textit{catastrophic forgetting} \cite{kirkpatrick2017overcoming}.
Despite many approaches exists to learn from data~\cite{dhar2021survey}, recently the focus has moved to improve the recognition accuracy of DL models because of their superior capabilities, accounting on new data belonging to known classes (\textit{domain-incremental CL}) or a new classes (\textit{class-incremental CL})~\cite{delange2021continual,mai2021online}. The CL techniques recently proposed are grouped in three categories: architectural, regularization and memory (or rehearsal) strategies. The architectural approaches specialize a subset of parameters for every (new and old) task but require the task-ID information at inference time, indicating the nature of current task in a multi-head network, and therefore they are not suitable for class or domain incremental continual learning. Concerning these latter scenarios, memory-based approaches, which preserve samples from previous tasks for replaying, perform better than regularization techniques, which simply address catastrophic forgetting by imposing constraints on the network parameter update at low memory cost~\cite{mai2021online,van2019three,chaudhry2019tiny}. This finding was confirmed during the recent CL competition at CVPR2020~\cite{lomonaco2020cvpr}, where the best entry leveraged on rehearsal based strategies. 

The main drawback of memory-based CL approaches concerns the high memory overhead for the storage of previous samples: the memory requirement can potentially grows over time preventing the applicability of these methods at the tiny scale, e.g.~\cite{mai2020batch}.
}
To address this problem, Pellegrini~et~al.~\cite{pellegrini2019latent} have recently introduced Continual Learning based on \textit{Latent Replays} (LRs).
The idea behind this is to combine a few \textit{old} data points taken from the original training set, but encoded intoa low-dimensional latent space to reduce the memory cost, with the new data for the incremental learning tasks. 
Hence, the previous knowledge is retained by means of Latent Replays samples, i.e. the intermediate feature maps of the DL model inference, selected so that they require less space with respect to the input data (up to 48$\times$ smaller compared to raw images~\cite{pellegrini2019latent}).
This strategy also leads to reduced computational cost: the Latent intermediate layer splits the network in a \textit{frozen} stage at the front and an \textit{adaptive} stage at the back, and only layers in the latter need to be updated.
So far, LR-based Continual Learning  has been successfully prototyped on high-performance embedded devices such as smartphones, including a Snapdragon-845 CPU running Android OS in the power envelope of a few Watts\footnote{https://hothardware.com/reviews/qualcomm-snapdragon-845-performance-benchmarks}. On the contrary, in this work, we focus on IoT applications and TinyML devices, with 100$\times$ tighter power constraints and 1000$\times$ smaller memories available.

In our preliminary work~\cite{ravaglia2020memory}, we proposed the early design concept of a HW/SW platform for Continual Learning based on the Parallel Ultra Low Power (PULP) paradigm \cite{rossi2015pulp}, and assessed the computational and memory costs to deploy Latent Replay-based CL algorithms.
{
 
In this paper, we complete and extend that effort by introducing several novel contributions from the software stack, system integration and algorithm viewpoint.
To the best of our knowledge, we present the first TinyML processing platform and framework capable of on-device CL, together with the design flow required to sustain learning tasks within a few tens of mW of power envelope ($>10\times$ lower than state-of-the-art solutions).
The proposed platform is based on \textit{VEGA}, a recently introduced end-node System-on-Chip prototype fabricated in 22nm technology \cite{rossi20214}. 
Unlike traditional low-power and flexible MCUs design, VEGA exploits explicit data parallelism, by featuring a multi-core SW programmable RISC-V cluster with shared Floating Point Units (FPUs),  DSP-oriented ISA and optimized memory management to enable the learning paradigm on low-end IoT devices.
}
Additionally, to gain minimum-cost on-device retention of Latent Replays and better enable deployment on an ultra-low-power platform, we extend the LR algorithm proposed by Pellegrini~et~al.~\cite{pellegrini2019latent} to work with a fully quantized \textit{frozen} front-end and compress Latent Replays using quantization down to 7 bits, with a small accuracy drop (almost lossless for 8-bit) when compared to the single-precision floating-point datatype (\textit{FP32}) on the Core50 CL classification benchmark.

% provide more details
In summary, the contributions of this work are:
\begin{enumerate}
\item We extend the LR algorithm to work with an 8-bit quantized and frozen front-end without impact on the CL process and to support LR compression with quantization, reducing up to 4.5$\times$ the memory needed for rehearsing. We call this extension \textit{Quantized Latent Replay-based Continual Learning} or \textit{QLR-CL}.
\item We propose a set of CL primitives including forward and backward propagation of common layers such as convolution, depthwise convolution, and fully connected layers, fine-tuned for optimized execution on VEGA, a TinyML platform for Deep Learning based on PULP \cite{rossi2015pulp}, fabricated in 22nm technology.
We also introduce a tiling scheme to manage data movement for the CL primitives.
\item We compare the performance of our CL primitives on VEGA with that on other devices that could in the future target on-chip at-edge learning, such as a state-of-the-art low-power STM32L4 microcontroller.
\end{enumerate}
Our results show that the Quantized Latent Replay based Continual Learning lead to a minimal accuracy loss on the Core50 dataset compared to the \textit{FP32} baseline, when compressing the Latent Replay memory by $4\times$ by means of 8-bit quantization. Compression to 7 bit can also be exploited but at the cost of a slightly lower accuracy, up to 5\% wrt the baseline when retraining one of the intermediate layer.
When testing the QLR-CL pipeline on the proposed VEGA platform, 
our CL primitives demonstrated to run up to $65 \times$ faster with respect to the MCUs for TinyML that can be found currently on the market. Compared against edge devices with a power envelope of 4W our solution is about $6 \times$ more energy-efficient, enough to operate 317h with a typical battery for embedded devices.
The rest of the paper is organized as follows: Section~\ref{sec:related} discusses related work in CL, inference and learning at the edge, and hardware architectures targeted at edge learning. Section~\ref{sec:methods} introduces the proposed methodology for Quantized Continual Learning. Section~\ref{sec:hwswcl} describes the HW/SW architecture of the proposed TinyML. Section~\ref{sec:experimental} evaluates and discusses experimental results. Section~\ref{sec:conclusion} concludes the paper.

\section{Related Work}
\label{sec:related}

\begin{table*}[t]
\begin{center}
    \caption{  On-device Learning methods on tiny embedded systems.}
    \begin{tabular}{ |l|l|c|c|c| c|c|c|c|} 
        \hline
        \textbf{Method} & \textbf{Learning} & \textbf{Problem} & \textbf{Proc.} & \textbf{Tiny} & \textbf{On-Device } & \textbf{Compute} & \textbf{Memory} & \textbf{Continual}\\ 
        \textbf{} & \textbf{Approach} & \textbf{} & \textbf{Device} & \textbf{Device} & \textbf{Learning} & \textbf{Cost} & \textbf{Cost} & \textbf{Learning}\\ 

        \hline
        \hline
        \textbf{Transfer} & Retraining last & Image & Coral &  & \checkmark & LOW & LOW & \\ 
        \textbf{Learning~\cite{cass2019taking}} & layer's weights & Classification & Edge TPU  & & & & & \\ 
        \hline

        \textbf{TinyTL} & Retraining Biases & Image & EPYC AMD & & \checkmark & MEDIUM & LOW / & \\ 
        \textbf{\cite{cai2020tinytl}} & & Classification & 7302 & & & & MEDIUM & \\ 
        \hline
        
        \textbf{TinyOL} & Add layer for  transfer-learning & Anomaly & Arduino Nano & \checkmark & \checkmark & LOW & LOW & \\
        \textbf{\cite{ren2021tinyol}} & based on streaming data & Detection & 33 BLE & & & & & \\ 
        \hline
        
        \textbf{TinyML} & CNN backprop. from scracth & Linear Camera  & GAP8 &  \checkmark & & - & - & \checkmark \\ 
        \textbf{Minicar~\cite{de2021robustifying}} & on increasing dataset & Class. 7 actions& & & & & & \\ 
        \hline
        
        \textbf{TML}  & kNN Classifier & Audio/Image & STM32F7 & \checkmark & \checkmark & LOW & HIGH & \checkmark\\ 
        \textbf{\cite{disabato2020incremental}} & & Class. 2 classes & & & &  & (unbounded) & \\ 
        \hline

        \textbf{PULP-HD} & Hyperdimensional & EMG 10 gestures & Mr. Wolf & \checkmark & \checkmark &  MEDIUM & LOW & \checkmark \\ 
        \textbf{\cite{benatti2019online}} & Computing & Classification  & & & & & & \\ 
        \hline

        \textbf{LR-CL} & CNN backprop. & Image & Qualcomm & & \checkmark & HIGH & HIGH / & \checkmark \\ 
        \textbf{\cite{pellegrini2019latent}} & w/ LRs & Class. 50 classes & Snapdragon  & & & & MEDIUM & \\ 
        \hline
        
        \textbf{QLR-CL} & CNN backprop. & Image & VEGA & \checkmark & \checkmark & HIGH & MEDIUM & \checkmark \\ 
        \textbf{[This Work]} & w/ Quantized LRs & Class. 50 classes  & & & & & & \\ 
        
        \hline
    \end{tabular}
    \label{tab:tiny_learn}    
\end{center}
\end{table*}

In this section, we first review the recent memory-efficient Continual Learning approaches before discussing the main solutions and methods for the TinyML ecosystem, including the first attempts for on-device learning on embedded systems.

\subsection{Memory-efficient Continual Learning}
Differently from \textit{Transfer Learning}~\cite{cass2019taking,cai2020tinytl}, which by design does not retain the knowledge of the primitive learned task when learning a new one,
\textit{Continual Learning} (CL)  has recently emerged as a new technique to tackle the acquisition of new/extended capabilities without losing the original ones -- a phenomenon known as \textit{catastrophic forgetting}~\cite{delange2021continual,mai2021online}.
One of the main causes of this phenomenon is that the newly acquired set breaks one of the main assumptions underlying supervised learning -- i.e., that training data are statistically independent and identically distributed (IID).
Instead, CL deals with training data that is organized in non-IID \textit{learning events}.
%to efficiently extend the previously static knowledge of the models to the ever-changing environment. 
Maltoni~et~al. in~\cite{maltoni2019continuous} sort the main CL techniques intothree groups: \textit{rehearsal}, which includes a periodic replay of the past information; \textit{architectural}, relying on a specialized architecture, layers, and activation functions to mitigate forgetting; and \textit{regularization}-based, where the loss term is extended to encourage retaining memory of pre-learned tasks.

Among these groups, \textit{rehearsal} CL strategies have emerged as the most effective to deal with catastrophic forgetting, at the cost of an additional replay memory~\cite{castro2018end, rebuffi2017icarl, pellegrini2019latent}.
In the recent CL challenge at CVPR2020 on the Core50 image dataset, $\sim$90\% of the competitors used rehearsal strategies~\cite{lomonaco2020cvpr}.
The best entry of the more challenging New Instances and Classes track (the same scenario considered in our work)~\cite{mai2020batch}, which is evaluated in terms of test accuracy but also memory and computation requirements, scores 91\% by replaying image data.
Unfortunately, this strategy results untractable for an IoT platform because of the expanding replay memory (up to 78k images) and the usage of a large DenseNet-161 model.
Conversely, the Latent Replay-based approach~\cite{pellegrini2019latent} relies on a fixed, and relatively small, amount of compressed \textit{latent} activations as replay data; it scores 71\% if retraining only the last layer, which presents a peak of $52\times$ lower (compressed) data points than the winning solution.
Additionally, the \textit{Jodelet} entry -- also employing LR-based CL -- achieves 83\% thanks to 3$\times$ more replays and a more accurate pre-trained model (ResNet50)~\cite{lomonaco2020cvpr}.
In our work, we focus on~\cite{pellegrini2019latent} because of the tunable accuracy-memory setting. Nevertheless, our proposed platform and compression methodology can be applied to any replay-based CL approach.

Also related to our work, ExStream~\cite{hayes2019memory} clusters in a streaming fashion the training samples before pushing them into the replay buffer while \cite{caccia2020online} uses discrete autoencoders to compress the input data for rehearsing. 
In contrast, we propose low-bitwidth quantization to compress the Latent Replay memory by $>$4$\times$ and, at the same time, reduce the inference latency and the memory requirement of the inference task of the \textit{frozen stage} if compared to a full-precision \textit{FP32} implementation.

\subsection{Deep Learning at the Extreme Edge}
Two main trends can be identified for TinyML platforms targeting the extreme edge.
On the one hand, Deep Learning applications are dominated by linear algebra which is an ideal target for application-specific HW acceleration~\cite{moonsEnvision26to10TOPSSubwordparallel2017,szeEfficientProcessingDeep2017}.
Most efforts in this direction employ a variety of inference-only acceleration techniques such as pruning~\cite{hanDeepCompressionCompressing2016} and byte and sub-byte integer quantization~\cite{choi2018pact}; the use of large arrays of simple MAC units~\cite{chenEyerissEnergyEfficientReconfigurable2017} or even mixed-signal techniques such as in-memory computing~\cite{legalloMixedprecisionInmemoryComputing2018}.

On the other hand, there are also many reasons for the alternative approach: running TinyML applications as software on top of commercial off-the-shelf (COTS) extreme-edge platforms, such as MCUs.
Extreme-edge TinyML devices need to be very cheap; they have to be flexible due both to economy of scale and to their need for integration within larger applications, composed of both neural and non-neural tasks~\cite{zemlyanikin512KiBRAMEnough2019}.
For these reasons, there is a strong push towards squeezing the maximal performance out of platforms based on COTS ARM Cortex-M class microcontrollers and DSPs, such as STMicroelectronics STM32 microcontrollers\footnote{\url{https://www.st.com/content/st_com/en/ecosystems/stm32-ann.html}}, or on multi-core parallel ultra-low-power (PULP) end-nodes, like GreenWaves Technologies GAP-8\footnote{\url{https://greenwaves-technologies.com/gap8_gap9/}}.
To cope with the severe constraints in terms of memory and maximum compute throughput of these platforms, a large number of deployment tools have been recently proposed.
Examples of this trend include non-vendor-locked tools such as Google TFLite Micro~\cite{david2020tensorflow}, ARM CMSIS-NN~\cite{lai2018cmsis}, Apache TVM~\cite{chenTVMEndtoendOptimization2018}, as well as frameworks that only support specific families of devices, such as STMicroelectronics~X-CUBE-AI\footnote{\url{https://www.st.com/en/embedded-software/x-cube-ai.html}}, GreenWaves Technologies NNTOOL\footnote{\url{https://greenwaves-technologies.com/sdk-manuals/nn_quick_start_guide}}, and DORY~\cite{burrello2021dory}.
Internally, these tools employ hardware-independent techniques, such as post-training compression \& quantization~\cite{capotondi2020cmix, cheng2017quantized, ghamari2021quantization}, as well as hardware-dependent ones such as data tiling~\cite{cecconiOptimalTilingStrategy2017} and loop unrolling to boost data reuse exploitation~\cite{lai2018cmsis}, coupled with automated generation of optimized backend code~\cite{moreau2018vta}.

As previously discussed, all of these efforts are mostly targeted at extreme edge inference, with little hardware and/or software dedicated to training.
Most of the techniques used to boost inference efficiency are not as effective for learning. For example, the vast majority of training is done in full precision floating-point (\textit{FP32}) or, with some restrictions, using half-precision floats (\textit{FP16})~\cite{kalamkar2019study} -- whereas inference is commonly pushed to \textit{INT8} or even below~\cite{capotondi2020cmix,choi2018pact}.
IBM has recently proposed a specialized 8-bit format for training called \textit{HFP8}~\cite{sunHybrid8bitFloating2019}, but its effectiveness is still under investigation. 

Hardware-accelerated on-device learning has so far been limited to high-performance embedded platforms (e.g., NVIDIA TensorCores on Tegra Xavier\footnote{https://www.nvidia.com/en-us/autonomous-machines/embedded-systems/jetson-xavier-nx} and mobile platforms such as Qualcomm Snapdragon 845~\cite{pellegrini2019latent}) or very narrow in scope.
For example, Shin~et~al.~\cite{shin201714} claim to implement an online adaptable architecture, but this is done using a simple  LUT to selectively activate parameters, and does not support more powerful mechanisms based on gradient descent.
A few recently proposed hardware accelerators for low-power training platforms~\cite{chen2014diannao,shin2020pragmatic,han2021hnpu,kang2021ganpu} enable partial gradient back-propagation by using selective and compressed weight updates, but they do not address the large memory footprint required by training.
Finally, several online-learning devices using bio-inspired algorithms such as Spiking Neural Networks~\cite{loboSpikingNeuralNetworks2020} and High-Dimensional Computing~\cite{benatti2019online} have been proposed~\cite{daviesLoihiNeuromorphicManycore2018,peiArtificialGeneralIntelligence2019,karunaratneRobustHighdimensionalMemoryaugmented2021}.
Most of these approaches, however, have only been demonstrated on simple MNIST-like tasks.

In this work, we propose the first, to the best of our knowledge, MCU-class hardware-software system capable of continual learning based on gradient back-propagation with a LR approach.
We achieve these results by leveraging on few key ideas in the state-of-the-art: \textit{INT8} inference, \textit{FP32} \textit{continual} learning, and exploitation of linear algebra kernels, back-propagation, and aggressive parallelization by deploying them on a multi-core FPU-enhanced PULP cluster.

{
 
\subsection{On-Device Learning on low-end platforms} 
Table~\ref{tab:tiny_learn} lists the main edge solutions featuring on-device learning capabilities. 
Every approach is evaluated by considering the memory and computational costs for the continual learning task and the suitability for deployment on highly resource-constrained (tiny) devices.

A first group of works deals with on-device transfer learning. The Coral Edge TPU, which presents a power budget of several Watts, features SW support for on-device fine-tuning of the parameters of the last fully-connected layer~\cite{cass2019taking}. TinyTL~\cite{cai2020tinytl} demonstrated on a high-end CPU that the transfer learning task results more effective (+32\% on the target Image Classification task) by retraining the bias terms and adding lite residual modules. 
TinyOL~\cite{ren2021tinyol} brought the transfer learning task on a tiny devices, i.e. an Arduino Nano platform featuring a 64MHz ARM Cortex-M4, by adding a trainable layer on top of a frozen inference model. Because only the coefficients of the last layer are updated during the online training process, no backpropagation of error gradients applies. 
Compared to these works, we address a continual learning scenario and therefore we provide a more capable and optimized HW/SW solution to match the memory and computational requirements of the adopted CL method.

Differently from the above works, \textit{de Prado} et al.~\cite{de2021robustifying} proposed a Continual Learning framework for self-driving mini-cars.
The embedded PULP-based MCU engine streams new data to a remote server, where the inference model is retrained from scratch on the enhanced dataset to improve the accuracy over time. This \textit{fully-rehearsal} methodology cannot be migrated to low-end devices because of the unconstrained increase of the memory footprint. 
In contrast, Disabato~et~al.~\cite{disabato2020incremental} presented an online adaptive scheme based on a kNN classifier placed on top of a frozen feature extraction CNN model. 
The final stage is updated by  incrementally adding the labeled samples to the knowledge memory of the kNN classifier. This approach has been evaluated on a tiny STM32F76ZI device but unfortunately has proven the effectiveness only on limited 2-classes problems and presents an unbounded memory requirement, which scales linearly with the number of training samples.
PULP-HD~\cite{benatti2019online} showed few-shot continual learning capabilities on an ultra-low power prototype using Hyperdimensional Computing. During the training phase the new data are mapped intoa limited hyperdimensional space by making use of a complex encoding procedure; at inference time the incoming samples are compared to the computed class prototypes. The method has been demonstrated on a 10 gesture classification scenario based on EMG data but lacks of experimental evidences to be effective on complex image classification problems.
In contrast to the these works, we demonstrate superior learning capabilities for a TinyML platform by
\textit{i)} running backpropagation on-device to update intermediate layers, and \textit{ii)} supporting a memory-efficient Latent Replay-based strategy to address catastrophic forgetting on a more complex Continual Learning scenario. 
An initial CNN-based prototype of a Continual Learning system was presented in in~\cite{pellegrini2019latent} using Latent Replays. The authors demonstrated the on-device learning capabilities using a Qualcomm Snapdragon processor, which features a power envelope 100$\times$ higher than our target and therefore it results not suitable for battery-operated tiny devices. In contrast to them, we also extend the LR algorithm by leveraging on quantization to compress the LR memory requirements.
}

\section{Methods}
\label{sec:methods}

In this section, we analyze the memory requirements of the Latent Replay-based Continual Learning method and present \textit{QLR-CL}, our strategy to reduce the memory footprint of the LR vectors based on a quantization process. 

\subsection{Background: Continual Learning with Latent Replays}
\label{sec:background}
In general, supervised learning aims at fitting an unknown function by using a set of known examples -- the training dataset.
In the case of Deep Neural Networks, the training procedure returns the values of the network parameters, such as weights and biases, that minimize a loss function. 
Among the used optimization strategies, the mini-batch Stochastic Gradient Descent (SGD), which is an iterative method applied over multiple learning step (i.e. the epochs),  is widely adopted. In particular, The SGD algorithm computes the gradient of the parameters based on the loss function by back-propagating the error value through the network.
This error function compares the model prediction, i.e. the output of the forward pass, with the expected outcome (the data label).
Parameter gradients obtained after the backward pass are weighted over a mini-batch of data before updating the model coefficients.

\begin{figure}[t]
    \centering
    \includegraphics[width=0.4\textwidth]{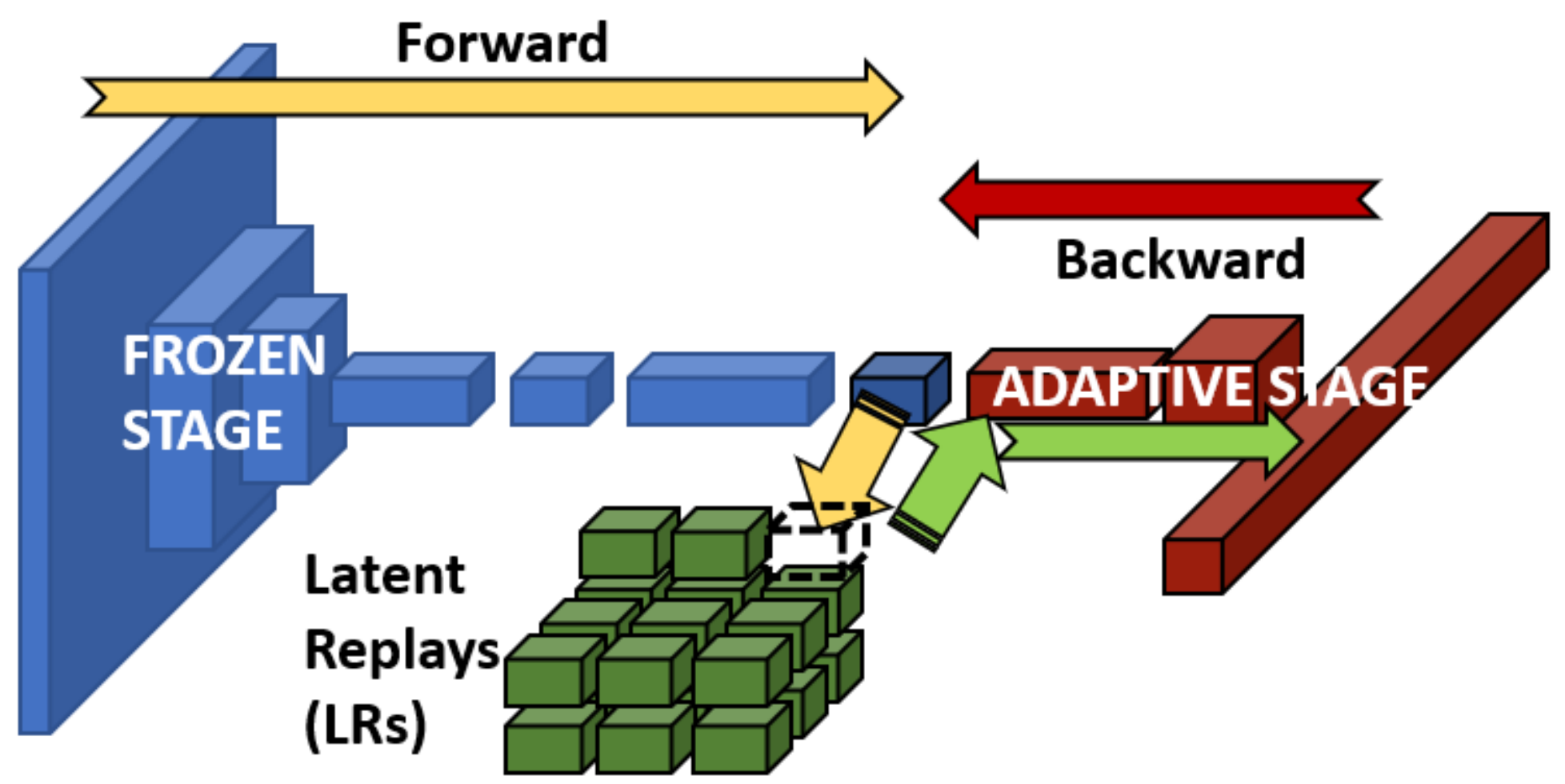}
    \caption{Continual Learning with Latent Replays. The frozen stage is the light-blue part (first half) of the network. After the first forward of the inputs (yellow arrow), activations (namely LRs) are stored apart. After having stored them, they will be used later mixed with the new images coming through the frozen stage and used to retrain the adaptive portion of the network.}
    \label{CLwLR}
\end{figure}

As introduced at the beginning of this work, the Latent Replay CL method~\cite{pellegrini2019latent} is a viable solution to gain TinyML adaptive systems with on-device learning capabilities based on the availability of new labeled data.
In Fig.~\ref{CLwLR} we illustrate the CL process with Latent Replays. 
The new data are injected into the model to obtain the latent embeddings, which are the feature maps of a specific intermediate layer.
We indicate such a layer with the index $l$, where $l\in[0,L)$, assuming the targeted model to be composed by $L$ stacked layers. 
At runtime, the new latent vectors are combined with the precomputed $N_{LR}$ Latent Replays vectors to execute the learning algorithm on the last $L-l-1$ layers. 
More specifically, the coefficient parameters of the \textit{adaptive stage} are updated by using a mini-batch gradient descend algorithm.
Every mini-batch includes both new data (in the latent embedding form) and LR vectors. The typical ratio of new data over the full mini-batch is 1/6~\cite{pellegrini2019latent}.
The coefficient gradients are computed through forward and backward passes over the adaptive (learned) layers.
Multiple iterations, i.e. the epochs, of the learning algorithms take place within the training procedure.

\label{sec:reducedLRCL}
\subsection{Memory Requirements} 
We model the Latent Replay-based Continual Learning task as operating on a set of new data coming from a sensor (e.g., a camera), which is interfaced with an embedded digital processing engine, namely the \textit{TinyML Platform}, and its memory subsystem.
Given the limited memory capacity of IoT end-nodes, the quantification of the learning algorithm's memory requirements is essential.
We distinguish between two different memory requirements: additional memory necessary for CL, e.g., the LR memory, and that required to save intermediate tensors during forward-prop to be used for back-prop -- a requirement common to all algorithms based on gradient descent, not specific to CL.

Concerning the LR memory, the system has to save a set of $N_{LR}$ LRs, each one of the size of the feature map computed at the  $l$-th layer of the network.
In our scenario, LR vectors are represented employing floating-point (\textit{FP32}) datatype and typically determine the majority of the memory requirement~\cite{ravaglia2020memory}.
Since LRs are part of the static \textit{long-term memory} of the CL system, for their storage, we use non-volatile memory, e.g., external Flash.

On the other hand, forward- and back-prop of the network model require to allocate the space for $N_P$ network parameters statically.
In addition, forward-prop requires dynamically allocated buffers to store the activation feature maps for all layers.
Up to the $l$-th layer, these buffers are temporary and can be released after their usage.
Conversely, the system must keep in memory the feature maps after $l$ to compute the gradients during back-prop.
They can only be released after the corresponding layer has been back-propagated.
Lastly, the system must also keep in memory the coefficients' gradients, demanding a second array of $N_P$ elements.
To keep accuracy on the learning process, every tensor, i.e. coefficients, gradients, and activations, employ a \textit{FP32} format in our baseline scenario.
Different from LRs, these tensors are kept intovolatile memories, except the frozen weights, which are stored in a non-volatile memory.

\begin{figure}[t]
    \centering
    \includegraphics[width=0.48\textwidth]{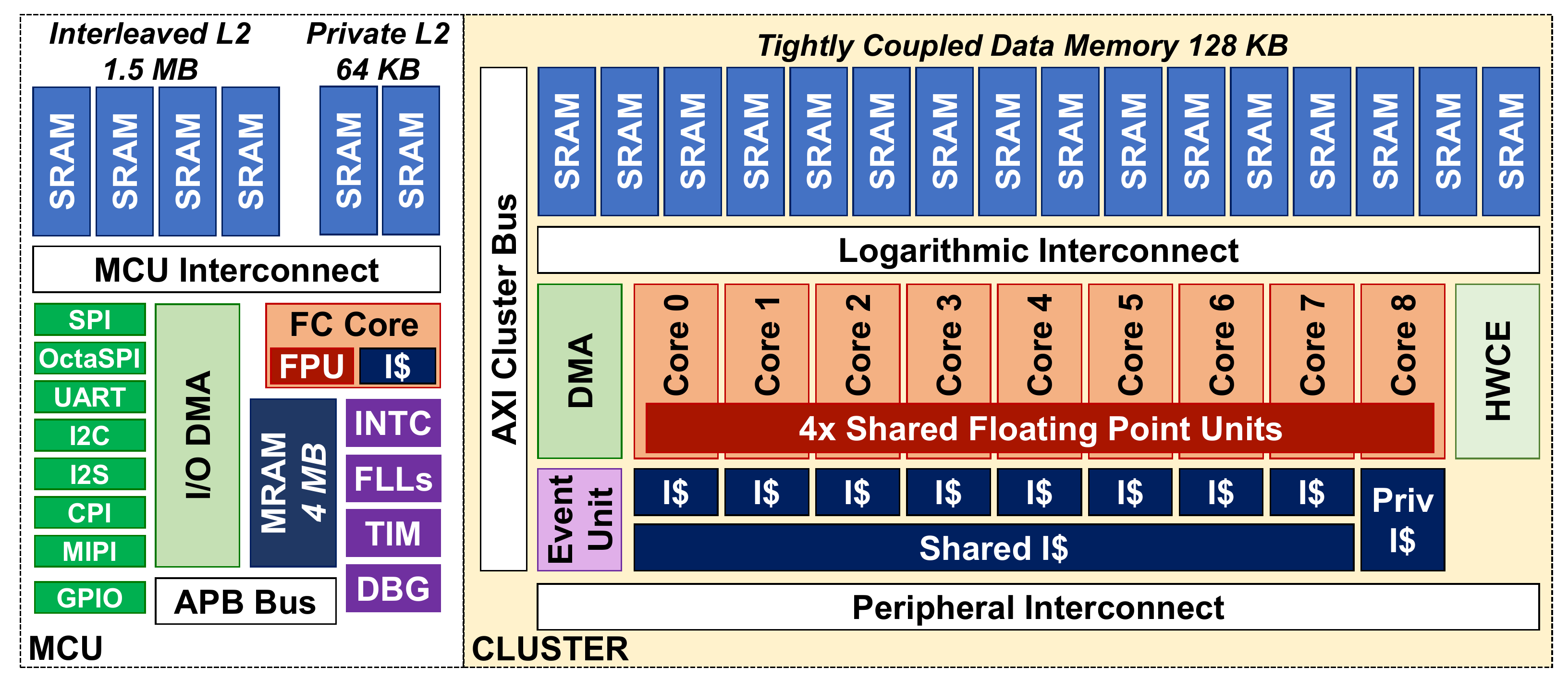}
    \caption{Architecture outline of the proposed PULP-based System-on-Chip for Continual Learning.}
    \label{fig:vega}
\end{figure}

% \subsection{CL Compression and LR Quantization} 
\subsection{Quantized Latent Replay-based Continual Learning}
\label{sec:compression}
Quantization techniques have been extensively used to reduce the data size of model parameters, and activation feature maps for the inference task, i.e. the forward pass. 
An effective quantization strategy reduces the data bitwidth from 32-bit (\textit{FP32}) to low bit-precision, 8-bit or less (\textit{Q} bits, in general) while paying an almost negligible accuracy loss.

In this paper, we introduce the Quantized Latent Replay-based Continual Learning method (QLR-CL) relying on low-bitwidth quantization to speed up the execution of the network up to the $l$-th layer and at the same time reduce the memory requirement of the LR vectors from the baseline \textit{FP32} arrays.
To do so, we split the deep model intotwo sub-networks, namely the \textit{frozen stage} and the \textit{adaptive stage}.
The \textit{frozen stage} includes the lower layers of the network, up to the Latent Replay layer $l$. The coefficients of this sub-network, including batch normalization statistics, are frozen during the incremental learning process. 
On the contrary, the parameters of the \textit{adaptive stage} are updated based on the new data samples. 

In QLR-CL, the Latent Replay vectors are generated by feeding the \textit{frozen stage} sub-network with a random subset of training samples from the CL dataset, which we denote as $X^{train}$.
The \textit{frozen stage} is initialized using pre-trained weights from a related problem -- in the case of Core50, we use a network pre-trained on the ImageNet-1k dataset.
Post-Training Quantization of the \textit{frozen stage} is based on training samples $X^{train}$.
We apply a standard Post-Training Quantization process that works by \textit{i)} determining the dynamic range of coefficient and activation tensors, \textit{ii)} dividing the range intoequal steps, using a uniform affine quantization scheme~\cite{jacob2018quantization}.
While the statistics of the parameters can be drawn without relying on data, the dynamic range of the activation features maps is estimated using $X^{train}$ as a calibration set.
If we denote the dynamic range of the weights at the $i$-th layer of the network as $[w_{i,min}, w_{i,max}]$, we can define the $w_{i,quant}$ \textit{INT-Q} representation of parameters as
\begin{equation}
    w_{i,quant} =\left\lfloor \frac{w_i}{S_{w,i}} \right\rceil, \quad S_{w,i} = \frac{ w_{i,max} - w_{i,min}}{2^Q -1}
    \label{eq:1}
\end{equation}
where $Q$ is the number of bits, $w_i$ is the full-precision output of the \textit{frozen stage}.
The representation of activations is similar, but we further restrict (\ref{eq:1}) for activations $a_i$ by considering the effect of ReLU's: $a_{i}$ are always positive and $a_{i,quant}$ can be represented using an unsigned \textit{UINT-Q} format:
\begin{equation}
    a_{i,quant} =\left\lfloor \frac{a_i}{S_{a,i}} \right\rceil, \quad S_{a,i} = \frac{ a_{i,max}}{2^Q -1}
    \label{eq:2}
\end{equation}
where $a_{i,max}$ is obtained through calibration on $X^{train}$.

Quantized Latent Replays (QLRs) $a_{l,replay}$ are represented similarly to other quantized activations, setting the layer $i$ to the LR $l$.
Their value is initialized during the initial setup of the QLR-CL process using the latent quantized activations $a_{l, quant}$ over the $X^{train}$ set.

During the QLR-CL process, the \textit{adaptive stage} is fed by dequantized vectors obtained as $S_{a,l} \cdot a_{l,replay}$, along with the dequantized latent representation of the new data sample $S_{a,l} \cdot a_{l,quant}$. Hence, the single \textit{FP32} parameter $S_{a,l}$ is also stored in memory as part of the \textit{frozen stage}. 
In our experiments, we set the bitwidth $Q$ of all activations and coefficients to 8-bit, while the output of the \textit{frozen stage} is compressed to 8-bit or less, as further explored in Section~\ref{sec:experimental}.

\section{Hardware/Software Platform}
\label{sec:hwswcl}

In this section, we describe the hardware architecture of the proposed platform for TinyML learning and the related software stack.

\subsection{Hardware architecture}
The CL platform we propose is inspired and extends on our previous work~\cite{ravaglia2020memory}.
We build it upon an advanced PULP-based SoC, called \textit{VEGA}, which combines parallel programming for high-performance with ultra-low-power features.
An advanced prototype of this platform has been taped out in GlobalFoundries 22nm technology~\cite{rossi20214}.
The system architecture, which is outlined in Fig.~\ref{fig:vega}, is based on an I/O-rich MCU platform coupled with a multi-core cluster of RISC-V ISA digital signal processing cores which are used to accelerate data-parallel machine learning \& linear algebra code.
The MCU side features a single RISC-V core, namely the Fabric-Controller (FC), and a large set of peripherals.
Besides the FC core, the MCU-side of the platform includes a large L2 SRAM, organized in an FC-private section of 64kB and a larger interleaved section of 1.5MB.
The interleaved L2 is shared between the FC core and an autonomous I/O DMA controller, connected to a broad set of peripherals such as OctaSPI/HyperBus to access an external Flash or DRAM of up to 64MB, as well as camera interfaces (CPI, MIPI) and standard MCU interfaces (SPI, UART, I2C, I2S, and GPIO).
The I/O DMA controller is connected to an on-chip magnetoresistive RAM (MRAM) of 4MB, which resides in its power and clock domain and can be accessed through the I/O DMA to move data to/from the L2 SRAM.

The multi-core cluster features nine processing elements (PE) that share data on a 128kB multi-banked L1 tightly coupled data memory (TCDM) through a 1-cycle latency logarithmic interconnect.
All cores are identical, using an in-order 4-stage architecture implementing the RISC-V \textit{RV32IMCFXpulpv2} ISA.
% The cluster can be configured to include private Floating Point Units (FPUs) per each core or a set of four FPU pipelines shared between all nine cores.
The cluster includes a set of four highly flexible FPUs shared between all nine cores, capable of \textit{FP32} and \textit{FP16} computation~\cite{mach2018transprecision}.
Eight cores are meant to execute primarily data-parallel code, and therefore they use a hierarchical Instruction cache (I\$) with a small private part (512B) plus 4kB of shared I\$~\cite{loi2017quest}.
The ninth core is meant to be used as a cluster controller for control-heavy data tiling \& marshaling operations; it has a private I\$ of 1kB.
The cluster also features a multi-channel DMA engine that autonomously handles data transfers between the shared L1 and the external memories through a 64-bit AXI4 cluster bus.
The DMA can transfer up to 8B/cycle between L2 and L1 TCDM in both directions simultaneously and perform 2D strided access on the L2 side by generating multiple AXI4 bursts.
The cluster can be switched on and off at runtime by the FC core employing clock-gating; it also resides on a separate power domain than the MCU, making it possible to completely turn it off and to tune its Vdd using an embedded DC-DC regulator.

\begin{figure}[t]
    \centering
    \includegraphics[width=0.95\linewidth]{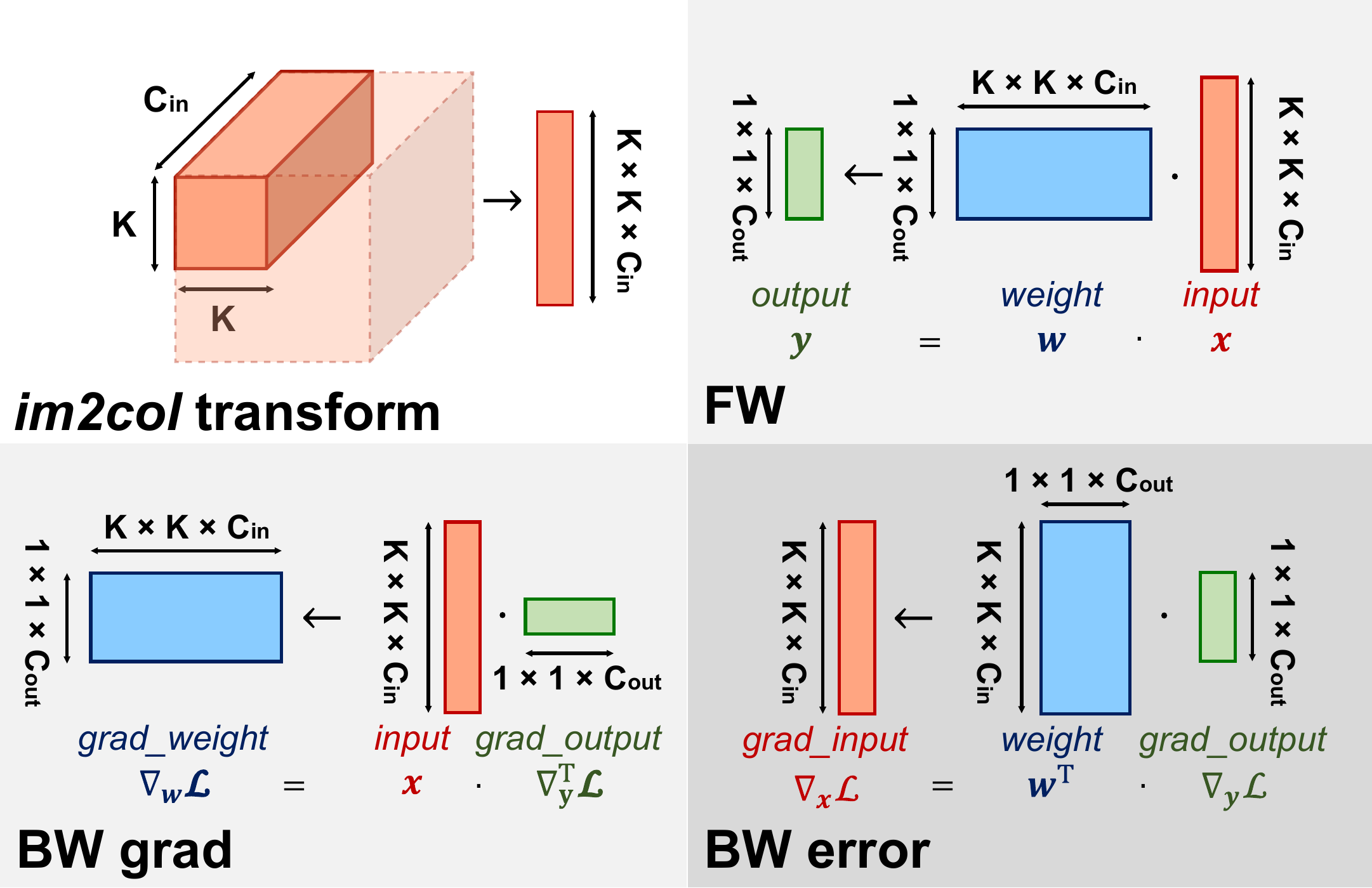}
    \caption{Clockwise from top-left: \textit{im2col} transform, Forward and Backward propagation for error and gradient calculation for a $K\times K$ Conv layer.}
    \label{fig:compute}
\end{figure}
\begin{figure}[t]
    \centering
    \includegraphics[width=0.8\linewidth]{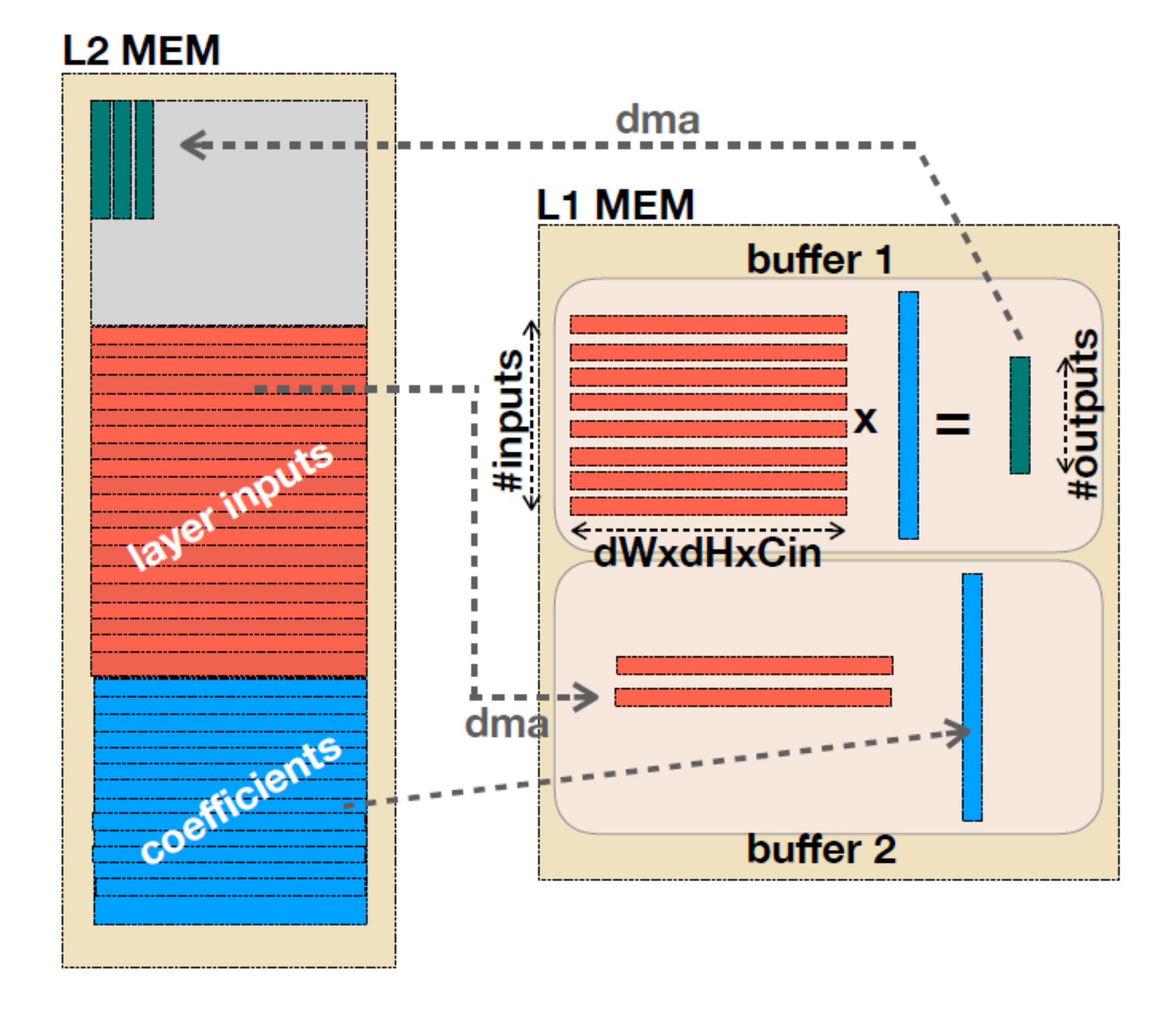}
    \caption{Tiling scheme between L2 and L1 memories exploiting double-buffering. 
    Two equal buffers are filled with the matrix multiplication terms that fit intohalf the size of L1. 
    The second buffer is filled with the next terms of the convolution that have to be matrix-multiplied.}
    \label{fig:tiling}
\end{figure}

\subsection{Software stack} 
\label{sec:software}
To execute the CL algorithm, the workload is largely dominated  by the execution of convolutional layers, such as pointwise, and depthwise, or fully connected layers ($\sim$98\% of operations in MobileNet-V1).
Consequently, the main load on computations is due to variants of matrix multiplications during the forward and backward steps, which can be efficiently parallelized on the 8 compute PEs of the cluster, leaving one core out to manage tiling and program data transfers.
Thus, to enable the learning paradigm on the PULP platform, we propose a SW stack composed of parallel layer-wise primitives that realize the \textit{forward step} and the back-propagation. 
The latter concerns either the computation of the activation gradients (\textit{backward error step}) and coefficient gradients (\textit{backward gradient step}).
Fig.~\ref{fig:compute} depicts the dataflow of the \textit{forward} and \textit{backward} for commonly used convolutional kernels such as pointwise (PW), depthwise (DW), and linear (L) layers.
To reshape all convolution operations intomatrix multiplications, the \textit{im2col} transformation is applied to the activation tensors to reshape them into2D matrix operands~\cite{lai2018cmsis}. The \textit{FP32} matrix multiplication kernel is parallelized over the eight cores of the cluster according to a data-parallelism strategy, making use of \textit{fmadd.s} (floating multiply-add) instructions made available by the shared FPU engines. 

The cores must operate on data from arrays located in the low-latency L1 TCDM to maximize throughput and computational efficiency (i.e., IPC). 
However, the operands of a layer function may not entirely fit into the lower memory level because of the limited space (128kB).
For instance, the tensors of the PW layer \#22 of the used MobileNet-V1 occupy 1.25MB.
Hence, the operands have to be sliced intoreduced-size blocks that can fit intothe available L1 memory and convolutional functions are applied on L1 tensor slices to increase the computational efficiency. 

This approach is generally referred to as \textit{tiling}~\cite{burrello2021dory}, which is schematized in Fig.~\ref{fig:tiling}.
By locating layer-wise data on the larger L2 memory (1.5MB), the DMA firstly copies individual slices of operand data, also referred to as \textit{tiles}, intoL1 buffers, to be later fetched by the cores.
Since the cluster DMA engine is capable of 2D-strided access on the L2 side, this operation can also be designed to perform \textit{im2col}, without any manual data marshaling overhead on L1.

To increase the computation efficiency, we implement a software flow that interleaves DMA transfers between L2 and L1 and calls to parallel primitives, e.g. \textit{forward}, \textit{backward error}, or \textit{backward gradient steps}, which operate on  individual tiles of data. Hence, every layer is expected to load and process all the tiles of any operand tensor. To reduce the overhead due to the data copy, the DMA transfers take place in the background of the multi-core computation: the copy of the next tile is launched before invoking the computation on loaded tiles. 
On the other side, this optimization requires doubling the L1 memory requirement: while one L1 buffer is used for computation, an equally-sized buffer is used by the data movement task. From a different viewpoint, the maximum tile size must not exceed half of the available memory.
At runtime, layer-wise \textit{tiled} kernels are invoked sequentially to run the learning algorithm with respect to the input data.
To this aim, LRs are loaded from external embedded memory, if not fitting the internal memory, and copied to the on-chip L2 memory thanks to the I/O DMA.
\section{Experimental Results}
\label{sec:experimental}
In this section, we provide the experimental evidence about our proposed TinyML platform for on-device Continual Learning. 
First, we evaluate the impact of quantization of the \textit{frozen stage} and the LR vectors upon the overall accuracy, and we analyze the memory-accuracy trade-off.
{
  
Secondly, we study the efficiency of the proposed SW architecture with respect to multiple HW configurations, namely \#cores, L1 size and DMA bandwidth, introducing the tiling requirements and evaluating the latency for each kernel of computation.} 
Then, we measure performance on an advanced PULP prototype, VEGA, fabricated in GlobalFoundries 22nm technology with 4 FPUs shared among all cores.
%We then validate these results on 
We analyze the latency results for individual layers forward and backward and estimate the overall energy consumption to perform a CL task on our platform.
Finally, we compare the efficiency of our TinyML platform to other devices used for on-device learning.

\subsection{Experimental Setup}
We benchmark the compression technique for the Latent Replay memory on the image-classification Core50 dataset, which includes 120k 128$\times$128 RGB images of 50 objects for the training and about 40k images for the testing. 
%To test the quantized version of the code, we use the Core50 dataset.
On the Core50 dataset, the CL setting is regulated by the NICv2-391 protocol~\cite{lomonaco2020rehearsal}.
According to this protocol, 3000 images belonging to ten classes are made available during the initial phase to fine-tune the targeted deep model on the Core50 problem. 
Afterward, the remaining 40 classes are introduced at training time in 390 learning events. 
Each event, as described more in detail in Section~\ref{sec:background}, comprises iterations over mini-batches of 128 samples each: 21 coming from actual images, all from the same class and typically not independent (e.g., coming from a video), and 107 latent replays.
After each learning event, the accuracy is measured on the test set, which includes samples from the complete set of classes. 

Following~\cite{pellegrini2019latent}, we use a MobileNet-V1 model with an input resolution of 128$\times$128 and width multiplier 1, pre-trained on ImageNet; we start from their public released code\footnote{Available at https://github.com/vlomonaco/ar1-pytorch/. 
While Pellegrini~et~al.~\cite{pellegrini2019latent} report lower accuracies in their paper, our \textit{FP32} baseline results are aligned with their released code.} and use PyTorch 1.5.
In our experiments, we replace BatchReNormalization with BatchNormalization layers and we freeze the statistics of the \textit{frozen stage} after fine-tuning.

\subsection{QLR-CL memory usage and accuracy}
\begin{figure}[t]
    \centering
    \includegraphics[width=\columnwidth]{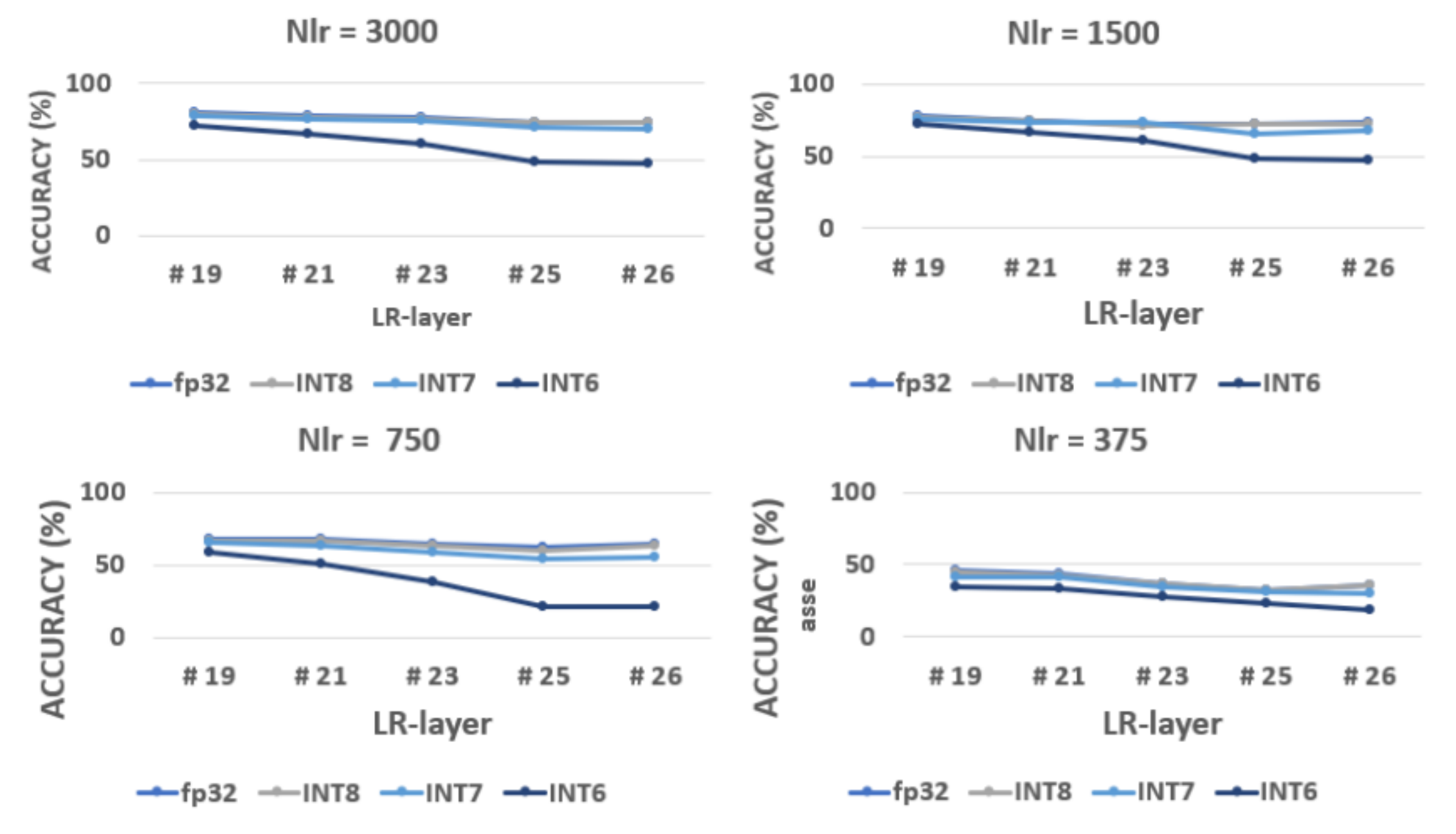}
    \caption{Accuracy plots for $N_{LR}=\{375,750,1500,3000\}$ and different levels of quantization. From these plots it is visible that below UINT-7 accuracy degrades rapidly.}
    \label{accuracy-precision}
\end{figure}

{
To evaluate the proposed \textit{QLR-CL} setting, we quantize the \textit{frozen} stage of the model using the PyTorch-based NEMO library~\cite{conti2020technical} after fine-tuning the MobileNet-V1 model with the initially available 3000 images.
We set the activation and parameters bitwidth of the \textit{frozen} stage to $Q=8$ bit while we vary the bitwidth $Q_{LR}$ of the latent replay layer. 
The quantized \textit{frozen} stage is used to generate a set of $N_{LR}$ Latent Replays, as sampled from the initial images.

The plots in Fig.~\ref{accuracy-precision} show the test accuracy on the Core50 that is achieved at the end of the NICv2-391 training protocol for a varying $N_{LR}=\{375,750,1500,3000\}$ while sweeping the LR layer $l$.
Depending on the selected layer type, the size of the LR vector varies as reported in Table~\ref{tab:activs}.

Each subplot of Fig.~\ref{accuracy-precision} compares the baseline \textit{FP32} version with our 8-bit fully-quantized solutions with a varying $Q_{LR}=\{8,7,6\}$, denoted in the figures, respectively, as UINT-8, UINT-7 and UINT-6. For a  $Q_{LR}<6$, we observe the Continual Learning process to not converge on the Core50 dataset. 

From the obtained results, we can observe the UINT-8 compressed solution featuring a small accuracy drop with respect to the full-precision \textit{FP32} baseline. When increasing the number of latent replays $N_{LR}$ to 3000, the UINT-8 quantized version results almost lossless (-0.26\%), if LR$=19$. On the contrary, if the LR layer is moved towards the last layer (LR=$27$), the accuracy drop increases up to 3.4\%. The same effect is observed when reducing $N_{LR}$ to 1500, 750 or 375. In particular, when $N_{LR}= 1500$, the UINT-8 quantatized version presents an accuracy drop from 1.2\% (LR$=19$) to 2.9\% (LR$=27$). 
%of the LR brings a very tiny accuracy loss compared to the \textit{FP32} baseline.
On the other hand, lowering the bit precision to UINT-7, the accuracy reduces on average of up to $5.2\%$, if compared to the \textit{FP32} baseline.
Bringing this further down to UINT-6 largely degrades the accuracy by more than 10\%.

To deeply investigate the impact of the quantization process on the overall accuracy, we perform an ablation study to distinguish the individual effects of \textit{i)} the quantization of the front-end and \textit{ii)} the quantization of the LRs. 
In case of $N_{LR} = 1500$, Table~\ref{tab:quant_cl} compares the accuracy on the Core50 dataset for different LR layers, if applying quantization to both the LR memory and the frozen stage or only to the LR memory. The accuracy statistics are averaged over 5 experiments; we report in the table the mean and the std deviation of the obtained results. 
In particular, we see that quantizing the LRs has a larger effect on the accuracy than quantizing the frozen graph. By quantizing only the LR memory to UINT-8, the accuracy drops by up to 1.2-2.6\% (higher in case of larger adaptive stages) with respect to the FP32 baseline. On the contrary, the UINT-8 quantized frozen graph brings only an additional 0.5-1\% of accuracy drop. With UINT-7 LRs, the accuracy drop is mainly due to the LR quantization: when compressing also the frozen stage to 8-bit the accuracy drop is up to -1\%, which is small compared to the total 4-7\% of accuracy degradation.

\begin{table}[t]
\caption{
Accuracy on Core50 dataset with multiple quantization settings \textbf{A+B}, where \textbf{A} denotes the quantization of the frozen stage (FP32 or UINT-8) and \textbf{B} indicates the quantization scheme of the LR vectors (FP32, UINT-8, UINT-7). The baseline is FP32. }

\resizebox{\columnwidth}{!}{
\begin{tabular}{|c||c||c|c||c|c|}
\hline
       \textbf{LR } & \textbf{FP32} & \textbf{FP32 +} & \textbf{UINT-8 +} & \textbf{FP32 +} & \textbf{UINT-8 +} \\
    layer & baseline & \textbf{UINT-8} & \textbf{UINT-8} & \textbf{UINT-7} & \textbf{UINT-7} \\
    \hline
    \hline
    \textbf{27} & \textbf{72.7} $\pm$ 0.34 & \textbf{70.1} $\pm$ 0.54 & \textbf{69.2} $\pm$ 0.48 & \textbf{68.0} $\pm$ 0.63 & \textbf{67.8} $\pm$ 1.14 \\
    \hline
    \textbf{25} & \textbf{73.3} $\pm$ 0.58 & \textbf{70.9} $\pm$ 0.65 & \textbf{70.2} $\pm$ 0.67 & \textbf{66.2} $\pm$ 0.75 & \textbf{66.1} $\pm$ 0.94 \\
    \hline
    \textbf{23} & \textbf{75.0} $\pm$ 0.83 & \textbf{73.2} $\pm$ 0.46 & \textbf{73.4} $\pm$ 0.66 & \textbf{71.1} $\pm$ 0.63 & \textbf{69.9} $\pm$ 1.25  \\
    \hline
    \textbf{21} & \textbf{76.5} $\pm$ 0.63 & \textbf{74.9} $\pm$ 0.51 & \textbf{73.9} $\pm$ 1.67 & \textbf{72.7} $\pm$ 0.74 & \textbf{72.6} $\pm$ 1.30 \\
    \hline
    \textbf{19} & \textbf{77.7} $\pm$ 0.73 & \textbf{76.5} $\pm$ 0.48 & \textbf{76.0} $\pm$ 0.80 & \textbf{74.0} $\pm$ 0.57 & \textbf{75.2} $\pm$ 1.10 \\
    \hline
\end{tabular}
}
\label{tab:quant_cl}

\end{table}

}

\begin{figure}[b]
    \centering
    \includegraphics[width=0.9\linewidth]{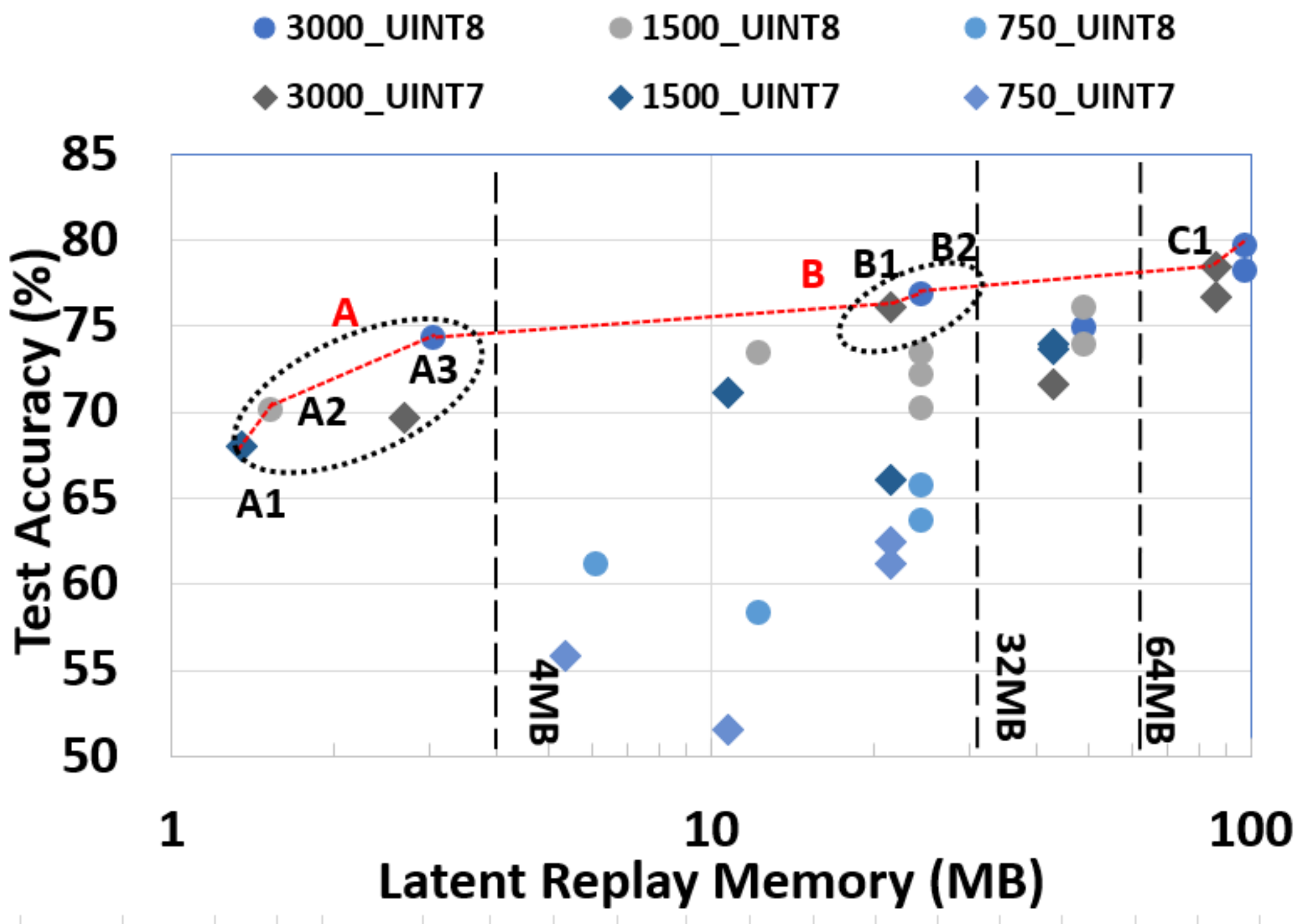}
    \caption{Accuracy achieved by considering $N_{LR}=\{750,1500,3000\}$ and different precision, highlighting the Pareto frontier.}
    \label{fig:accuracy-mem-pareto}
\end{figure}

To facilitate the interpretation of the results, Fig.~\ref{fig:accuracy-mem-pareto} reports the test accuracy for multiple quantization settings compared to the size (in MB) of the Latent Replay Memory.
%The closest point to the top-left corner, \textbf{A2}, has $l=27$ with 1500 LR UINT-8, and is the ``optimal'' point if we consider the error-delay product as our metric.
In red, we highlight a Pareto frontier of non-dominated points, to have a range of options to maximize accuracy and minimize the memory footprint.
Among the best solutions, we detect two clusters of points on the frontier.
The first cluster (\textbf{A}), corresponding to the low-memory side of the frontier, is constituted by experiments that use $l=27$ with 1500 or 3000 LRs and UINT-7 or UINT-8 representation.
On the other hand, if we aim at the highest accuracy possible for our QLR-CL classification algorithm, we can follow the Pareto frontier to the right towards higher accuracies at steeper memory cost, reaching cluster \textbf{B}.
All points in cluster \textbf{B} features $l=23$ as Latent Replay layer, which is a bottleneck layer of the network and allows to store more compact tensors as LR (refer to Table~\ref{tab:activs}). 
Adopting LR layers within \textbf{B} leads accuracy to an average of $76\%$, gaining $\sim$5$\%$ on average with respect to the layers within cluster \textbf{A}.
A single point \textbf{C1} is shown further to the right, but still below 128MB.
%-- the maximum size that would fit within a typical Flash for TinyML devices.

\begin{figure}[t]
    \centering
    \includegraphics[width=0.4\textwidth]{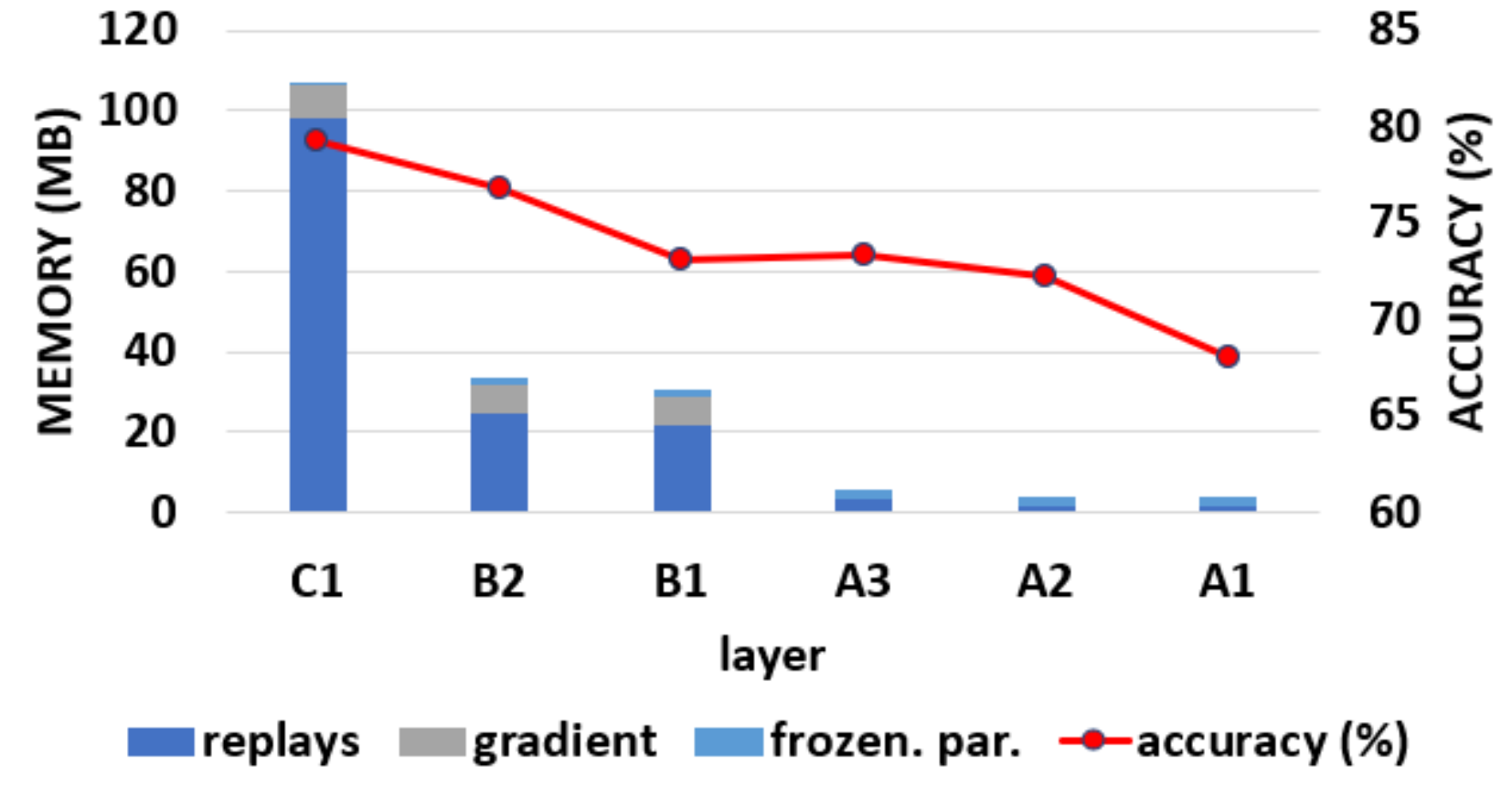}
    \caption{Memory requirements for the points highlighted in Fig.~\ref{fig:accuracy-mem-pareto}. Each layer belongs to the Pareto frontier and accounts for all the memory components. Going deeper into the network, LRs (gray) dominate memory consumption. The other components are the  parameters of the frozen stage, the gradient and the activations needed during the training.}
    \label{columns}
\end{figure}

For a deeper analysis of the Pareto frontier, in Fig.~\ref{columns}, we detail the memory requirements when analyzing the points into the two clusters \textbf{A} and \textbf{B}, as well as \textbf{C1}. 
We make two observations: first, in all \textbf{A} points, it would be possible to fit entirely within the on-chip memory available on VEGA, exploiting the 4MB of non-volatile MRAM.
This would allow avoiding any external memory access, increasing the energy efficiency of the algorithm by a factor of up to $\sim$3$\times$~\cite{rossi20214}.
Moreover, considering that the maximization of accuracy is often the primary objective in CL, we observe that accumulating features at $l=19$ with 1500 UINT-8 LRs (point \textbf{C1}) enables accuracy to grow above $77\%$, almost $10\%$ more than the compact solutions in \textbf{A} (Fig. \ref{columns}).
This analysis allows us to also speculate over possible future architectural explorations to design optimized bottleneck layers that could facilitate better memory accuracy trade-off for QLR-CL.

\begin{table}[t]
\begin{center}
    \caption{Size of tiles for the MobileNet-V1 layers.}
    \begin{tabular}{ |c||c|c|c|c|c|c|c| } 
        \hline
        \textbf{LR} & \textbf{Layer}  & \textbf{LR Dim.} & \textbf{LR Size} \\
        \textbf{Layer $l$}  & \textbf{Type}  & \textbf{($H\times W\times C$)}   & (\#elements) \\
        \hline
        \hline
        19    & DW & 8$\times$8$\times$512 &  32k  \\ 
        \hline
        20    & PW & 8$\times$8$\times$512 &  32k  \\
        \hline
        21    & DW & 8$\times$8$\times$512 &  32k  \\ 
        \hline
        22    & PW & 8$\times$8$\times$512 &  32k  \\
        \hline
        23    & DW & 4$\times$4$\times$512 &  8k   \\ 
        \hline
        24    & PW & 4$\times$4$\times$1024 &  16k \\ 
        \hline
        25    & DW & 4$\times$4$\times$1024 &  16k \\ 
        \hline
        26    & PW & 4$\times$4$\times$1024 &  16k \\
        \hline
        27    & Linear    & 1$\times$1$\times$1024 &  1k \\
        \hline
    \end{tabular}
    \label{tab:activs}    
\end{center}
\end{table}

\begin{figure*}[]
    \centering
    \includegraphics[width=0.95\textwidth]{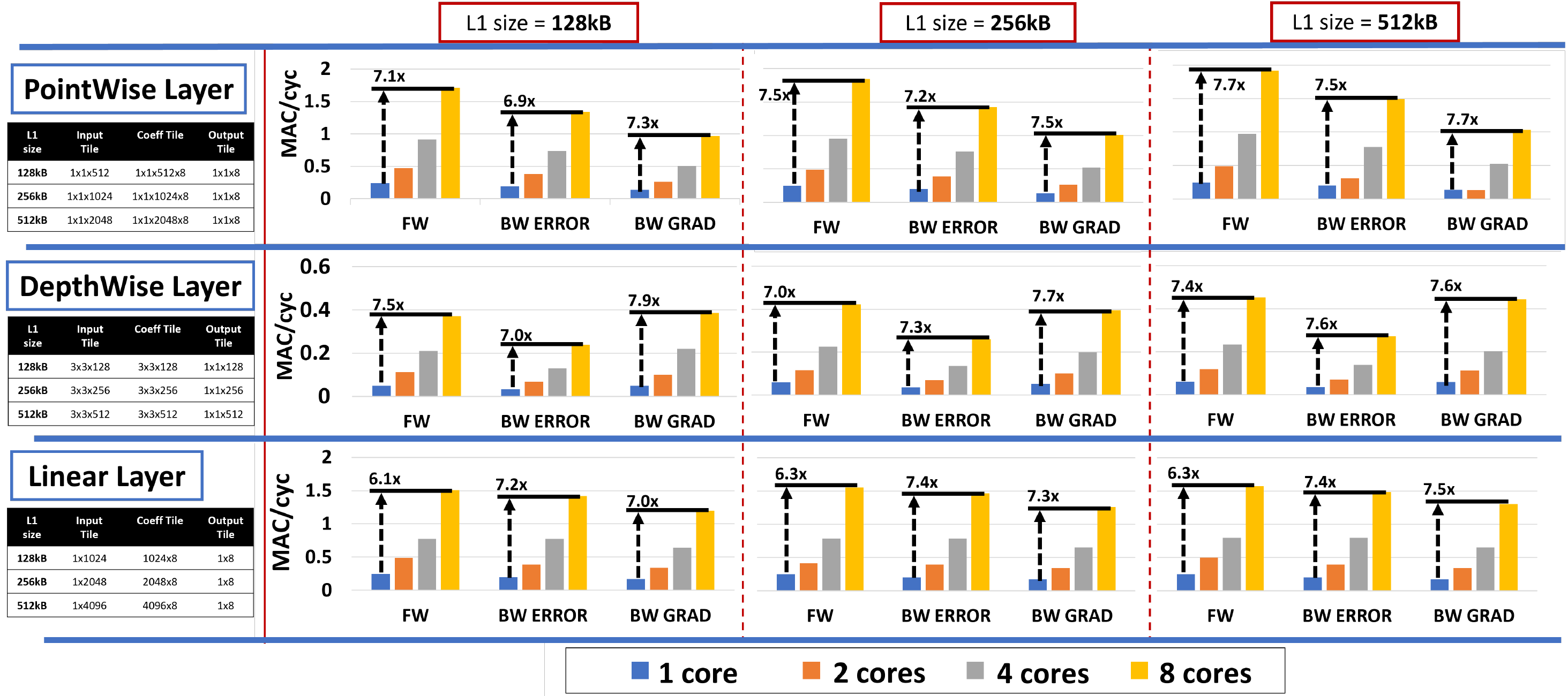}
    \caption{  
    Efficiency, expressed in MAC/cyc, of the proposed CL primitives for forward and backward pass: PointWise,  DepthWise, and Linear layers. The analysis concerns a varying number of cores (1, 2, 4 or 8) and L1 memory size (128, 256 or 512 kB), which impacts on the dimension of the layer's tensor tiles as reported in the tables on the left. } 
    \label{fig:mac-cycle}
\end{figure*}

{

\subsection{Hardware/Software Efficiency}
To assess the performance of the proposed solution, we study the efficiency of the CL Software primitives on the target platform and the sensitivity to some of the HW architectural parameters, namely the \#cores, the L1 memory size and the cluster DMA Bandwidth. 

\subsubsection*{Single-tile performance on L1 TCDM}
Based on the tiling strategy described in Section~\ref{sec:software}, we run experiments concerning the CL primitives of the software stack that operates on individual tiles of data placed in the L1 memory.
Figure~\ref{fig:mac-cycle} shows the latency performance, expressed as \textit{MAC/cyc}, i.e. the ratio between Multiply-Accumulate operations (\textit{MAC}) and elapsed clock cycles (\textit{cyc}), for each of the main \textit{FP32} computation kernels in case of single-core (\textit{1-CORE}) or multi-core (\textit{2-4-8-CORES}) execution. 
We highlight that a higher value of MAC/cyc denotes a more efficient processing scheme, leading to lower latency for a given computation workload, i.e. fixed MAC.
More specifically, in this plot, we evaluate the forward (\textit{FW}), backward error (\textit{BW ERR}), and backward gradient (\textit{BW GRAD}) for each of the considered layer for a varying size of the L1 TCDM memory, i.e. 128, 256 or 512kB.
The shapes of the tiles for PointWise (\textit{PW}), DepthWise (\textit{DW}), and Linear (\textit{Lin}) layers used for the experiments are reported in the tables on the left of the figure. Such dimensions are defined to fit three different sizes of the TCDM, considering buffers of size 64kB, 128kB and 256kB.

Focusing firstly on the \textit{PW} layers (histograms at the top of the figure), we observe a peak performance in the 8-cores \textit{FW} step, achieving  up to 1.91 MAC/cyc for a L1 memory size of 512kB. 
We observe also a performance improvement  of up to 11\%  by increasing the L1 size from 128kB to 512kB, which is is motivated by the higher computational density of the kernel: if L1$=512$kB the inner loop features 4$\times$  iterations than a scenario with 128kB of L1 size. 
Moreover, the parallel speedup scales almost linearly with respect to the number of cores and archives 7.2$\times$ in case of 8 cores. 
With respect to the theoretical maximum of 8$\times$, the parallel implementation presents some overheads mainly due to increased L1 TCDM contentions and cluster’s cache misses.  

If we look at \textit{DW} convolutions, their performance is lower with respect to the others. The main reason is that it requires a software-based \textit{im2col} data layout transformation, which increase the amount of data marshaling operations and adds an extra L1 buffer, thus reducing the size of matrices in the matrix multiplication, leading to increased overheads. Specifically, we measure the workload of the \textit{im2col} to achieve up to 70\% of the FW kernel's latency.
As mentioned in Section~\ref{sec:hwswcl}, the primitives we introduce also support performing the im2col directly when moving the data tile from L2 via DMA transfer – in that case, this source of performance loss is not present, and the MAC/cyc necessary for depthwise convolutions increases up to 1 MAC/cycles for depthwise forward-prop, depending also on the L1 size selected.
The remaining overhead with respect to pointwise convolutions is justified by the fact that  depthwise convolutions can only exploit filter reuse (of size 3$\times$3, for example, in MobileNet-V1 DW layers) and no input channel data-reuse, resulting in much shorter inner loops and more visible effect of overheads.
This latter effect cannot be counteracted by efficient DMA usage; on the other hand, since depthwise convolutions account for less than $1.5\%$ of the computation, their impact on the overall latency is limited, as we further explore in the following section.

Moving our analysis towards the different performance between forward- and backward-prop layers (particularly BW grad), we observe that this effect is again due to different data re-use between the matrix multiplication kernels. 
The reduction in re-use in the backward-prop is due to the tiling strategy adopted (see Fig.~\ref{fig:compute}) has a \textit{grad\_output} vector which is shorter than the input in the forward matrix multiplication. 
Specifically, the input to the matrix multiplication has size 8x1x1 in backward, while the input shape in forward changes accordingly with the L1 memory: 512x1x1 for 128kB L1, 1024x1x1 for 256kB L1 and 2048 for 512kB L1. 
In this scenario, the inner loop of the matrix multiplication of a forward computation is 64$\times$, 128$\times$ or 256$\times$ larger with respect to the backward kernels' cases. 
This fact motivates the lower MAC/cyc of the \textit{BW ERR} step (22\%) and \textit{BW GRAD} step (-46\%) if compared to the FW kernel.
}

\begin{figure}[]
    \centering
    \includegraphics[width=\columnwidth]{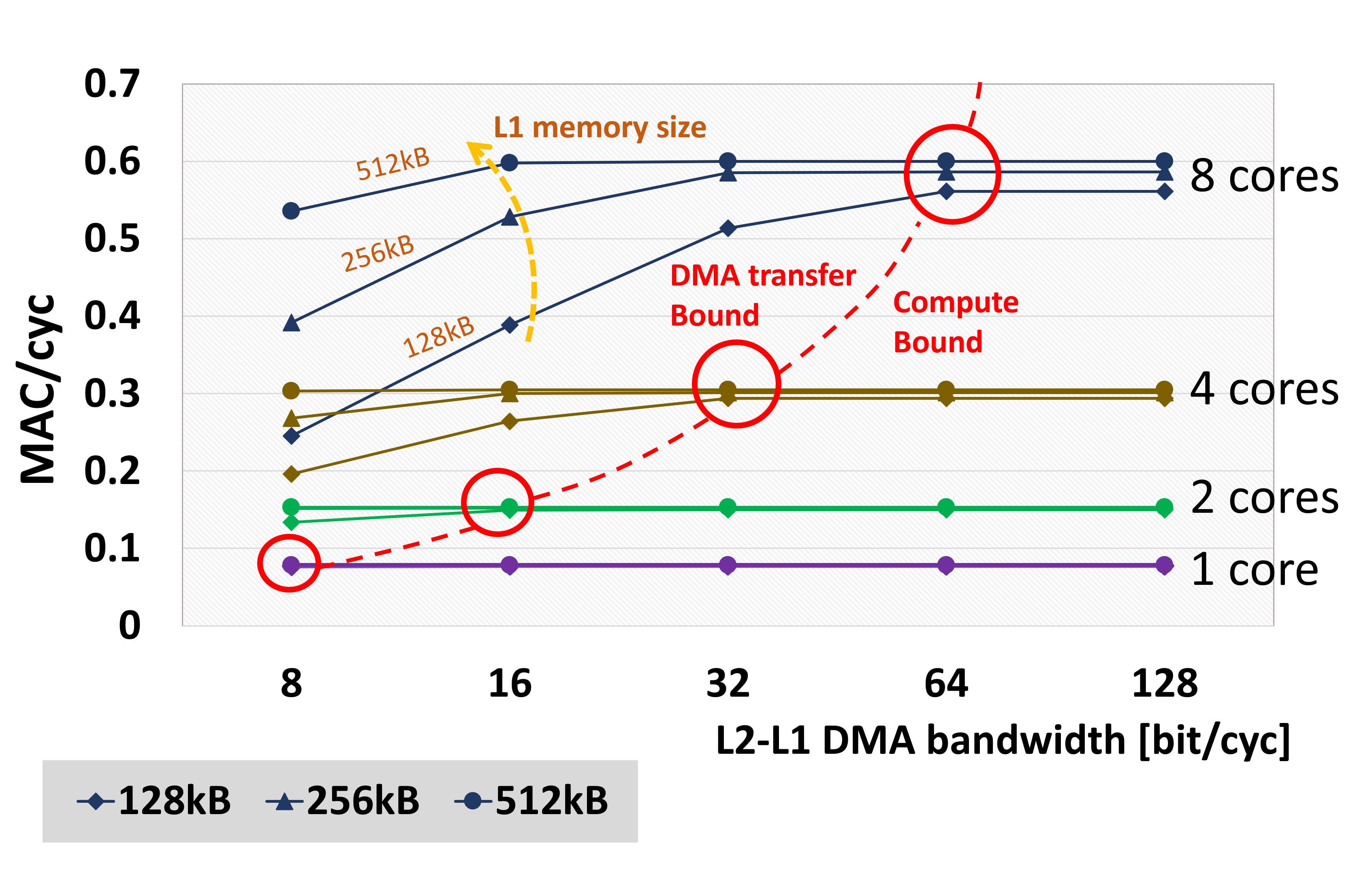}
    \caption{  
    SW efficiency, expressed as average MAC/cyc, when running forward and backward steps with respect to a varying L1-L2 bandwidth. Every line corresponds to a configuration of \#cores (1, 2, 4 or 8 cores) and L1 memory size (128, 256 or 512kB).
    }
    \label{fig:mac-cycle-global}
\end{figure}

{
 
\subsubsection*{L2-L1 DMA Bandwidth effects on performance}
Next we analyze the impact of  L2-L1 DMA Bandwidth variations, due to the Cluster DMA, on the overall performance of the learning task. In particular, we monitor the latency and the MAC/cyc for multiple values of L2-L1 bandwidth ranging from 8 to 128 bits per clock cycle (bit/cyc) and different configurations of \#cores and L1 size. We remark that a higher value of MAC/cyc indicates a better performing HW configuration. Our analysis assumes a single half-duplex DMA channel, hence the bandwidth value accounts for either read or write transfers. 

Fig.~\ref{fig:mac-cycle-global} reports the average MAC/cyc when running the forward and backward steps with respect to the L2-L1 cluster's DMA bandwidth. 
%with respect to a single new sample on 
As a benchmark, we consider the adaptive stage of the MobileNetV1 model when the LR layer is set to the 19th layer.
Hence, we adopt our tiling strategy and double-buffering scheme to realize the training.
When increasing the L1 size, the tensor tiles feature a larger size, therefore demanding a higher transfer time to copy data between the L1 memory (used for computation) and L2 memory (used for storage). 
Thanks to the adopted double-buffering technique, such transfer time can be hidden by the computation time because the DMA works in the background of CPU operation (\textit{compute-bound}). On the contrary, if the transfer time results dominating, the computation becomes \textit{DMA transfer-bound}, with lower benefits from the multi-core acceleration.

In case of single core execution, the measured MAC/cyc does not vary with respect to the L1 size (128kB, 256kB or 512kB) as can be seen from the plot. In this scenario, the CPU time results as the dominant contribution with respect to the transfer time: the execution is compute-bound and a higher L2-L1 bandwidth does not impact the overall performance. Differently, in a multi-core execution (2, 4 or 8 cores), the average MAC/cyc increases and therefore the ratio between transfer time and the computation time decreases: from the plot we can observe higher performance if the DMA bandwidth is increased. If featuring a L1 size of 128kB, the sweet spots between DMA and compute bound are observed when the L2-L1 DMA bandwidth is 16 (2 cores), 32 (4 cores) and 64 (8 cores) bit/cyc, respectively, as highlighted by the red circles in the plot. These configurations denote the sweet spots to tune the DMA requirements with respect to the chosen L1 memory size and \#cores. 

If focusing more on the impact of the L1 memory size to the multi-core performance, we observe up to 2$\times$ efficiency gain with 8 cores with a larger L1 memory, increasing from 0.25 MAC/cyc for a 128kB L1 memory to 0.4MAC/cyc at L1=256kB and to 0.53MAC/cyc for 512kB of L1. At 64 bit/cyc of L2-L1 DMA bandwidth, the execution, which is dominated by the computation, reaches 0.52MAC/cyc, 2.12$\times$ faster than the low-bandwidth configuration.

From this analysis we can conclude that the best design point for the learning task on a low-end multi-core architecture can be pinpointed leveraging the L2-L1 DMA Bandwidth and the L1 memory size tuning: when using 8 cores, 128kB of L1 memory, which is typically the main expensive resource for the system, can lead already to the highest performance as long as the DMA features a bandwidth of 64 bit/cyc. On the contrary, if the DMA's bandwidth is as low as 8 bit/cyc, a 512 kB L1 memory is needed to gain maximum performance. The target chip VEGA includes a L1 memory of 128 kB; the DMA follows a full-duplex scheme and can provide up to 64 bit/cyc for read transactions and 64 bit/cyc for write transactions. Therefore the VEGA HW architecture can fully exploit the presented SW architecture and optimization schemes to reach the optimal utilization and performance for the learning task.
}

\begin{table}[t]
\begin{center}
\caption{Cumulative latency values per learning event for VEGA, STM32, and Snapdragon845.
}
\resizebox{\columnwidth}{!}{%
\begin{tabular}{|c||*{6}{c|}}
\hline
\textbf{LR} & \multicolumn{3}{c|}{\textbf{VEGA} @ 375 MHz} & \multicolumn{2}{c|}{\textbf{STM32L4} @ 80 MHz} & \multicolumn{1}{c|}{\textbf{Snapdragon}}\\
% \cline{2-5}
\textbf{Layer} & \textbf{Adaptive}  & \textbf{Frozen} & \textbf{Cumul.} & \textbf{Total} & \textbf{Cumul.} &  \textbf{Total}\\
\textbf{$l$} & \textbf{[s]}  & \textbf{[s]} & \textbf{En. [J]} & \textbf{[s]} & \textbf{En. [J]} &  \textbf{[s]}\\

\hline
\hline
20 & $2.49 \cdot 10^{3}$  & 0.87  & 154  & $1.65 \cdot 10^{5}$ & 5688 & n.a. \\
\hline
21 & $1.73 \cdot 10^{3}$   & 0.94 & 107  & $1.15 \cdot 10^{5}$ & 3981 & n.a. \\
\hline
22 & $1.64 \cdot 10^{3}$   & 0.95 & 101 & $1.08 \cdot 10^{5}$ & 3728 & n.a. \\
\hline
\textbf{23} & $8.77 \cdot 10^{2}$ & 1.03  & 54.3 & $5.86 \cdot 10^{4}$ & 2020 & n.a. \\
\hline
24 & $7.81 \cdot 10^{2}$   & 1.03  & 48.4 & $5.12 \cdot 10^{4}$ & 1769 & n.a. \\
\hline
25 & $4.01 \cdot 10^{2}$   & 1.09 & 24.9 & $2.65 \cdot 10^{4}$ & 915 & n.a. \\
\hline
26 & $3.81 \cdot 10^{2}$ & 1.10 & 23.5 & $2.49 \cdot 10^{4}$ & 859 & n.a. \\
\hline
\textbf{27} & 2.07 & 1.25  & 0.13 & $1.39 \cdot 10^{2}$ & 4.80 & 0.50 \\
\hline
 \end{tabular}
 }
\label{latency-table}
\end{center}
\end{table}

\subsection{Latency Evaluation on VEGA SoC}

We run experiments on the VEGA SoC to assess the on-device learning performance, in terms of latency and energy consumption, of the proposed QLR-CL framework. Specifically, we report the computation time, i.e. the latency, at the running frequency of 375MHz and the power consumption by measuring the current absorbed by the chip when powered at 1.8V. 
To measure the full layer latency, we profile \textit{forward} and \textit{backward} tiled kernels, which include DMA transfers of data, initially stored in L2, and calls to low-level kernel primitives, introduced above. 
%We measure the latency and power consumption when running the tiled kernels on VEGA.
On average, we observe a 7\% of tiling overhead with respect to the single-tile execution on L1. 
This is not surprising, due to the large bandwidth availability between L1 and L2 and the presence of compute-bound matrix multiplication operations.

Based on the implemented tiled functions, we report the layer-wise performance in Table~\ref{latency-table} for any of the layers of the MobileNet-V1 model.
We consider as complete time for the execution of a layer the cumulated time for \textit{frozen stage} and \textit{adaptive stage}. 
The latency of the \textit{frozen stage} is obtained using DORY~\cite{burrello2021dory} to deploy the network, as this operation is performed as pure 8-bit quantized inference.
We compute the full latency of the \textit{adaptive stage} as the time needed to execute the forward and backward phases of each layer.
Since we have multiple configurations, latencies for retraining start growing from the last layer (\#27) up to layer \#20, where retraining comprises a total of eight layers. 

First of all, we note that \textit{frozen stage} latencies are utterly dominated by the \textit{adaptive stage}.
Apart from the faster inference backend, which can rely on 8-bit SIMD vectorization, this is because only 21 images per mini-batch pass through the \textit{frozen stage}, while the \textit{adaptive stage} has to be executed on 128 latent inputs (107 LRs and the 21 dequantized outputs from the \textit{frozen stage}), and it has to run for multiple epochs (by default, 4) in order to work.

When $l=27$, the \textit{adaptive stage} is very fast thanks to its very small number of parameters (it produces just the 50 output classes).
This is the only case in which the frozen stage is non-negligible ($\sim$ 1/6 of the overall time).
Progressing upward in the table, the frozen stage becomes negligible.
The cumulative impact of forward and backward passes through all the other layers we take intoaccount ($l$ from \#20 to \#26) is in the range between 0.3h and 1.5h.
In particular, $l=23$ corresponds to $\sim$14 min per learning event;  this LR layer corresponds to high accuracy ($>$75\% in Core50, see Fig.~\ref{fig:accuracy-mem-pareto}), which means that in this time the proposed system is capable of acquiring a very significant new capability (e.g., a new task/object to classify) while retaining previous knowledge to a high degree.

Having the basic mini-batch measurements, we can estimate any scenario, by considering that to train with 1500 LR and $l=27$, we will need 300 new images, thus we need 14 mini-batches ($300/21$), which leads to 3.30 seconds to learn a new set of images, with an accuracy of 69.2\%.
If we push back the LR layer $l$, this leads to an increase of accuracy 76.5\%, at the expense of much larger latency, up to 42 minutes for layer \#20 (see Table~\ref{latency-table}).

\subsection{Energy Evaluation on CL Use-Cases and Comparison with other Solutions}

To understand the performance of our system and its real-world applicability, we study two use-cases: a single mini-batch of the Core50 training we used, and the simplified scenario presented by Pellegrini~et~al.~\cite{pellegrini2019latent} in their demonstration video.
We compare our results with another MCU targeting ultra-power consumption: a NUCLEO-64 board based on the STM32L476RG MCU, on which we ran a direct port of the same code we use on the PULP-based platforms.
It has two on-chip SRAMs with 1-cycle access time and an overall capacity of 96kB.
Performance results, in terms of latency, are reported in Table~\ref{latency-table}, where we take intoaccount the cumulative latency values both for VEGA and STM32 implementations, along with the cumulative energy consumption.
Cumulative latency is computed by adding from the linear layer of the network the latencies of the preceding layers.
{
 
On average, execution on VEGA's 8-cores on performs \textbf{65$\times$} faster with respect to the STM32 solution thanks to three main factors. Firstly, the clock frequency of VEGA is 4.7$\times$ higher than the max clock frequency of the STM32L4 (375MHz vs 80MHz), also thanks to the superior technology node. Secondly, VEGA presents a parallel speed-up of up to 7.2$\times$. Lastly, thanks to the more optimized ISA and the core microarchitecture, VEGA performs less operations while executing the same learning task. For example, the inner loop of the matrix multiplication on VEGA requires only 4 instructions while the STM32L4 takes 9 instructions, resulting 2.25$\times$ faster, mainly thanks to the HW loop extension and the fmadd.s instruction.

The latency speed up, leads to an energy gain of around \textbf{37$\times$}, because the average power consumption of VEGA is \textbf{2$\times$} higher than the STM32L4 at full load.

Notice that the latency measurement of the STM32L4 does not account for possible overheads due to the tiling data between the small on-chip SRAM banks and off-chip memory.
Even then, our results show that fine-tuning from any layer above the last one results in too large a latency to be realistic on STM32L4 -- in the order of a day {per learning event} with $l=23$. 
On the contrary, CL on VEGA can be completed in $\sim 14$ minutes if selecting $l=23$ or as fast as 3.3 seconds if retraining only the last layer.

\begin{figure}[tb]
    \centering
    \includegraphics[width=0.5\textwidth]{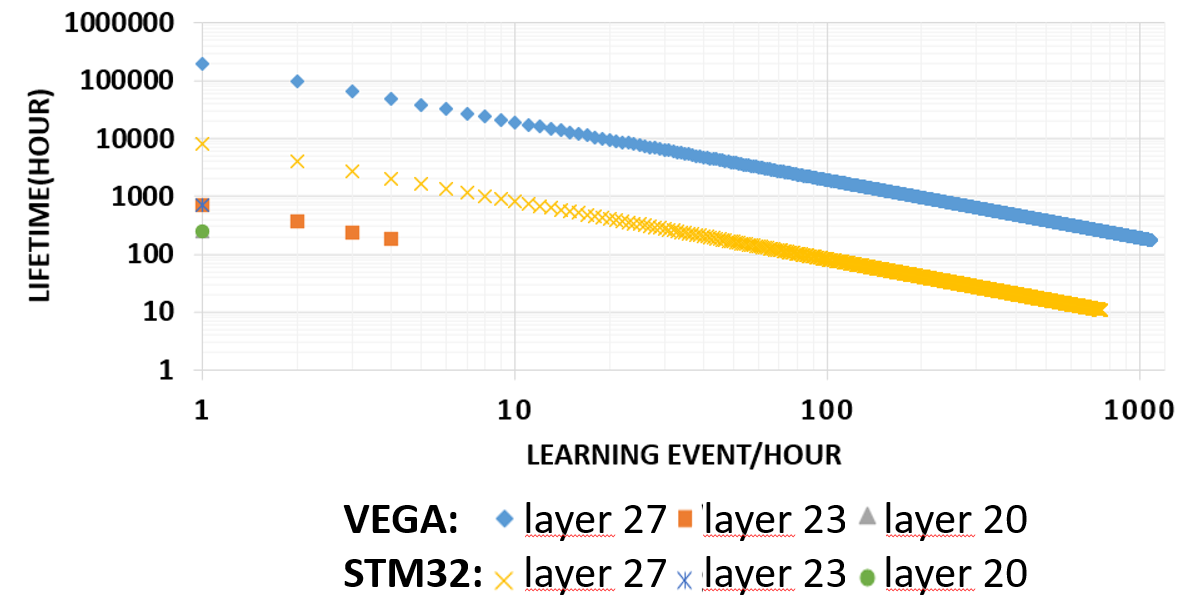}
    \caption{  
    Battery Lifetime of the VEGA SoC and the STM32L4 devices when handling multiple learning events per hour. 
    }
    \label{fig:lifetime-retraing}
\end{figure}
}

Given the reported energy consumption, we estimated the battery lifetime of our device when adapting the inference model by means of multiple learning events per hour; we assumed no extra energy consumption for the remaining time. 
In particular, Fig.~\ref{fig:lifetime-retraing} shows the battery lifetime (in hours) depending on the selected Latent Replay layer and the adaptation rate, expressed as the amount of learning events per hour.
We considered a 3300 mAh battery as the only energy source for the device.
By retraining only the last layer (LR$=27$), an intelligent node featuring our device can perform more than 1080 continual learning events per hour, leading to a lifetime of about 175h. On the contrary, if retraining larger portions of the network, the training time increases and the maximum rate of the learning events reduces to less than 10/hour, with a lifetime in the range 200-1000h. In comparison, on a STM32L4, if retraining the coefficients of the last layer, the maximum learning rate per hour is limited to 750, with a lifetime of about 10h. This latter can be increased up to 10000h but retraining only once in one hour. At the same learning event rate, the battery lifetime of VEGA is 20x higher.

Lastly, we compare with the use-case presented by Pellegrini~et~al.~\cite{pellegrini2019latent}, where they developed a mobile phone application that performs CL with LRs on a OnePlus6 with Snapdragon845. 
For this scenario, they consider only 500 LRs before the linear layer, these will be shuffled with 100 new images.
Then, by construction the mini-batch is composed of 100 LRs and 20 new images, thus, for each of the 8 training epochs, the network will process 5 times over the 20 new images and the 100 LRs.
This scenario leads them to obtain an average latency of 502 ms for a single learning event. 
On the other hand, considering our measurements on VEGA we obtain a forward latency of 1.25s and a training time of 2.07s for a whole learning event.

Considering the power envelope of a Snapdragon845 of about 4W, and the average power consumption of VEGA of 62mW, this implies that our solution is 9.7$\times$ more efficient in terms of energy. 
We additionally assess the energy consumption and the duration of a battery in the mobile application scenario, provided the energy measurements on VEGA, when using a 3300mAh battery.
Thus, if we consider performing learning over a mini-batch of images once every minute in the ultra-fast scenario (just retraining the linear layer) and to perform an inference each second, we obtain an energy consumption of 0.25J per minute. This leads the accuracy of the model to achieve an average of 69.2\%, with an overall lifetime of about 108 days.

\section{Conclusion}
\label{sec:conclusion}

In this work, we presented what, to the best of our knowledge, is the first HW/SW platform for TinyML Continual Learning -- together with the novel Quantized Latent Replay-based Continual Learning (QLR-CL) methodology.
More specifically, we propose to use low-bitwidth quantization to reduce the high memory requirements of a Continual Learning strategy based on Latent Replay rehearsing. 
We show a small accuracy drop as small as 0.26\% if using 8-bit quantized LR memory if compared to floating-point vectors and an average degradation of 5\% if lowering the bit precision to 7-bit, depending on the LR layer selected. 
Our results demonstrate that sophisticated adaptive behavior based on CL is within reach for next-generation TinyML devices, such as PULP devices; we show the capability to learn a new Core50 class with accuracy up to 77\%, using less than 64MB of memory -- a typical constraint to fit Flash memories.
We show that our QLR-CL library based on VEGA achieves up to $\sim$65$\times$ better performance than a conventional STM32 microcontroller.

These results constitute an initial step towards moving the TinyML from a strict \textit{train-then-deploy} approach to a more flexible and adaptive scenario, where low power devices are capable to learn and adapt to changing tasks and conditions directly in the field.
{
 
Despite this work focused on a single CL method, we remark that, thanks to the flexibility of the proposed platform, other adaptation methods or models can be also supported, especially if relying on the back-propagation algorithm and  CNN primitives, such as convolution operations.
}
\section*{Acknowledgement}
We thank Vincenzo Lomonaco and Lorenzo Pellegrini for the insightful discussions.

\bibliographystyle{IEEEtran}

\bibliography{bibliography}

\begin{IEEEbiography}[{\includegraphics[width=1in,height=1.25in,clip,keepaspectratio]{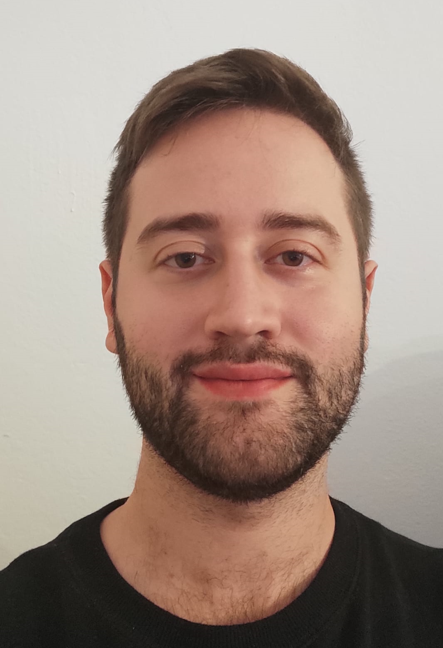}}]%
{Leonardo Ravaglia} receive his M.Sc. degree in Automation Engineering from the University of Bologna in 2019. 
He is currently a Doctoral Student in Data Science and Computation at the University of Bologna. 
His research interests include DNN algorithms for Continual Learning, parallel computing on Ultra Low Power devices and  Quantized Neural Networks.
\end{IEEEbiography}
\vskip 0pt plus -1fil
\begin{IEEEbiography}[{\includegraphics[width=1in,height=1.25in,clip,keepaspectratio]{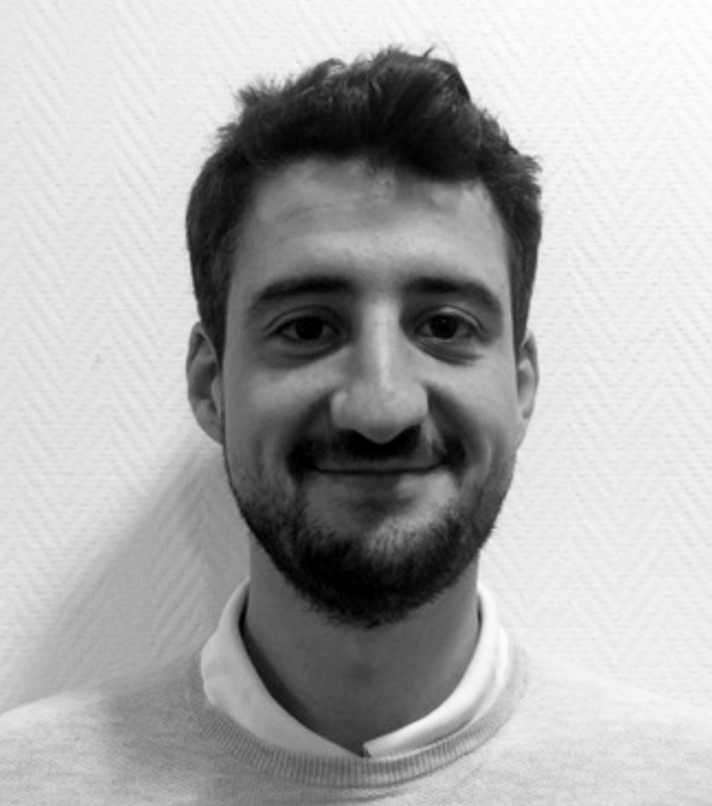}}]%
{Dr. Manuele Rusci} received the Ph.D. degree in electronic engineering from the University of Bologna in 2018. 
He is currently a Post-Doctoral Researcher at the same University at the Department of Electrical, Electronic 
and Information Engineering “Guglielmo Marconi” and closely collaborates with Greenwaves Technologies. 
His main research interests include low-power embedded systems and AI-powered smart sensors. 
\end{IEEEbiography}
\vskip 0pt plus -1fil
\begin{IEEEbiography}[{\includegraphics[width=1in,height=1.25in,clip,keepaspectratio]{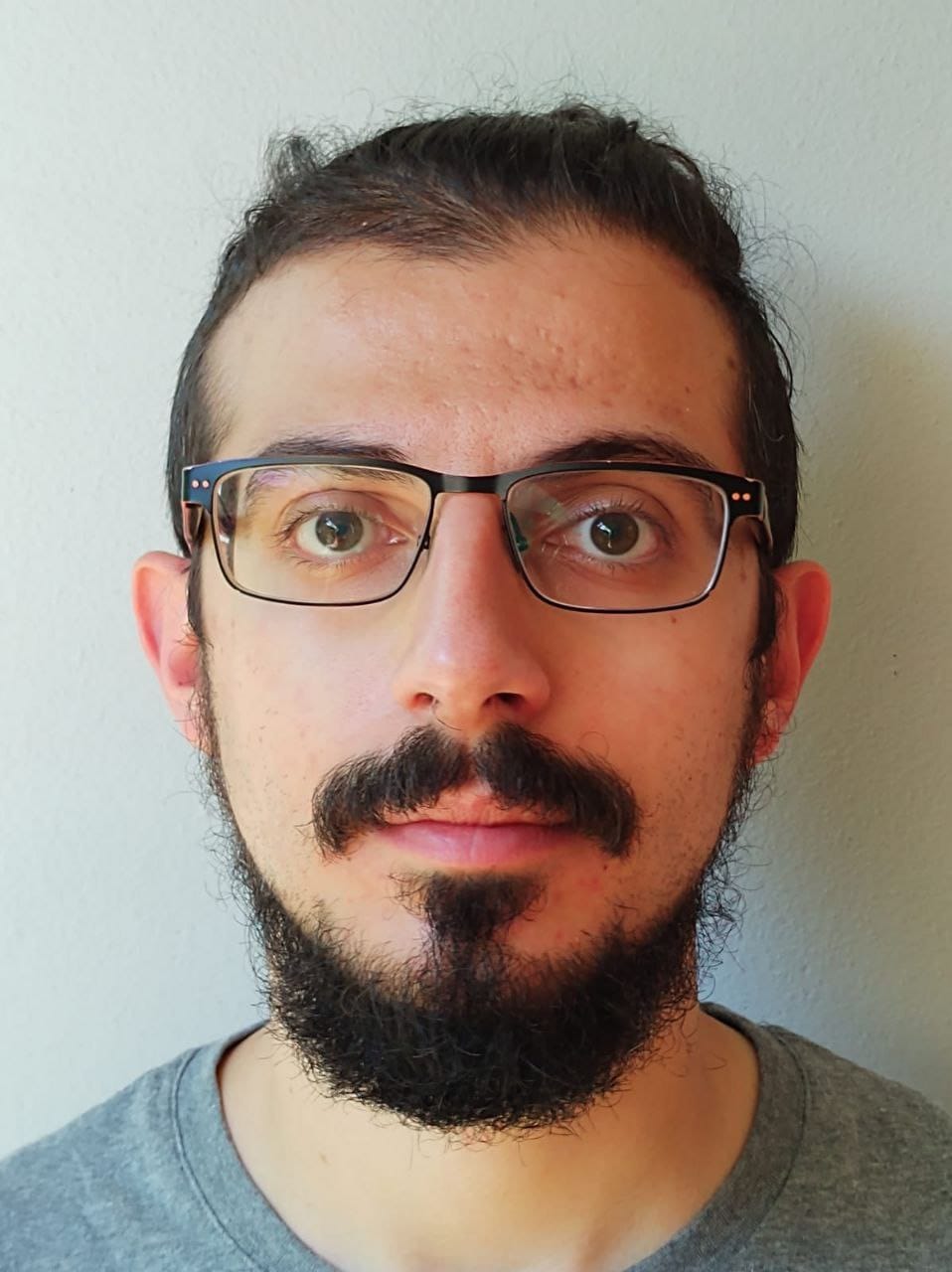}}]%
{Davide Nadalini} Davide Nadalini received the M.Sc. degree in electronic engineering from the University of Bologna in 2021. Since then, he is a Ph.D. student at Politecnico di Torino. His main research topic is Hardware-Software co-design and optimization of embedded Artificial Intelligence. His research interests include parallel computing, Quantized Neural Networks and low-level optimization. 
\end{IEEEbiography}
\vskip 0pt plus -1fil
\begin{IEEEbiography}[{\includegraphics[width=1in,height=1.25in,clip,keepaspectratio]{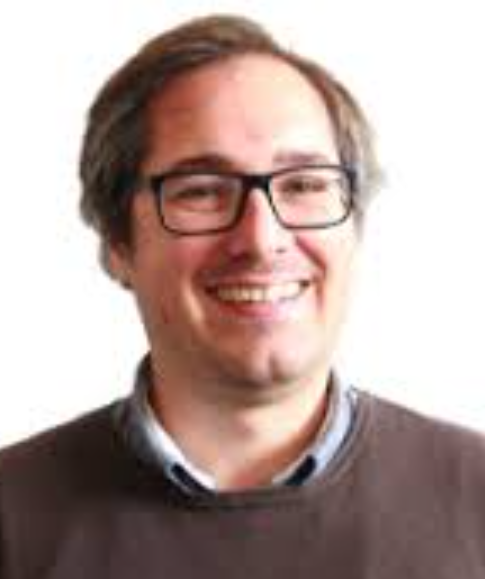}}]%
{Dr. Alessandro Capotondi} Dr. Alessandro Capotondi (IEEE Member) is a postdoctoral researcher at the Universit\`a di Modena e Reggio Emilia (IT).
His main research interests focus on heterogeneous architectures, parallel programming models, and TinyML.
He received his Ph.D. in Electrical, Electronic, and Information Engineering from the University of Bologna in 2016.
\end{IEEEbiography}
\vskip 0pt plus -1fil
\begin{IEEEbiography}[{\includegraphics[width=1in,height=1.25in,clip,keepaspectratio]{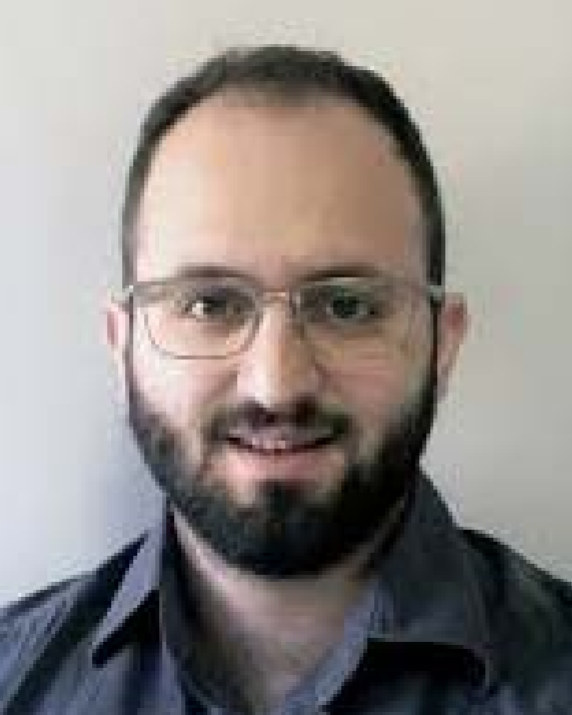}}]%
{Prof. Francesco Conti} received the Ph.D. in electronic engineering from the University of Bologna, Italy, in 2016.
He is currently an Assistant Professor in the DEI Department of the University of Bologna.
From 2016 to 2020, he was a postdoctoral researcher at the Integrated Systems Laboratory of ETH Zürich in the Digital Systems group.
His research is focused on the development of deep learning based intelligence on top of ultra-low power, ultra-energy efficient programmable Systems-on-Chip. In particular, he works on Deep Learning-aware architecture, on tinyML hardware acceleration facilities such as dedicated accelerator cores and ISA extensions, as well as on automated DNN architecture search, quantization, and deployment methodologies tuned to maximally exploit hardware features.
He has been involved in the development of the RISC-V based open-source Parallel Ultra-Low-Power (PULP) project since its inception (2013).
From 2020, he has collaborated with GreenWaves Technologies, France as a consultant for the development of DNN and RNN acceleration IP.
His research has resulted in 50+ publications in international conferences and journals and has been awarded several times, including the 2020 IEEE TCAS-I Darlington Best Paper Award, the 2018 Hipeac Tech Transfer Award, the 2018 ESWEEK Best Paper Award, and the 2014 ASAP Best Paper Award.
\end{IEEEbiography}
\vskip 0pt plus -1fil
\begin{IEEEbiography}[{\includegraphics[width=1in,height=1.25in,clip,keepaspectratio]{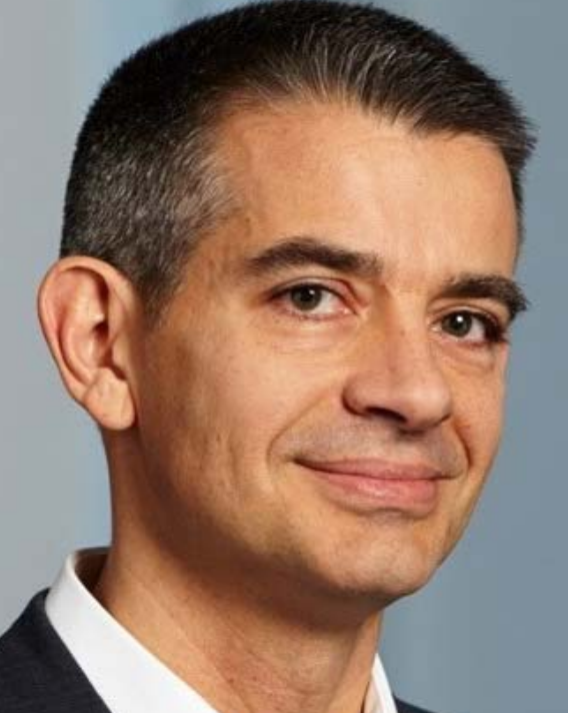}}]%
{Prof. Luca Benini} (Fellow, IEEE) received the Ph.D.
degree in electrical engineering from Stanford
University, Stanford, CA, USA, in 1997.
He was the Chief Architect of the Platform
2012/STHORM Project with STMicroelectronics,
Grenoble, France, from 2009 to 2013. He held
visiting/consulting positions with École Polytechnique Fédérale de Lausanne, Lausanne, Switzerland;
Stanford University; and IMEC, Leuven, Belgium.
He is currently a Full Professor with the University
of Bologna, Bologna, Italy. He is also the Chair
of Digital Circuits and Systems with ETH Zürich, Zürich, Switzerland.
He has authored over 700 papers in peer-reviewed international journals and
conferences, four books, and several book chapters. His current research
interests include energy-efficient system design and multicore system-on-chip
design.
Dr. Benini is a member of Academia Europaea. 
\end{IEEEbiography}

\end{document}